\documentclass[journal]{IEEEtran}
\usepackage{amssymb}
\usepackage{multirow}
\usepackage{color}

\usepackage{cite}
\ifCLASSINFOpdf
  \usepackage[pdftex]{graphicx}
  \graphicspath{{../pdf/}{../jpeg/}}
  \DeclareGraphicsExtensions{.pdf,.jpeg,.png}
\else
  \usepackage[dvips]{graphicx}
  \graphicspath{{../eps/}}
  \DeclareGraphicsExtensions{.eps}
\fi
\usepackage{amsmath}
\usepackage{algorithm}  
\usepackage{algorithmic} 
\usepackage{array}
\begin{document}
%
% paper title
% Titles are generally capitalized except for words such as a, an, and, as,
% at, but, by, for, in, nor, of, on, or, the, to and up, which are usually
% not capitalized unless they are the first or last word of the title.
% Linebreaks \\ can be used within to get better formatting as desired.
% Do not put math or special symbols in the title.
\title{1D-Convolutional Capsule Network\\for Hyperspectral Image Classification}
%
%
% author names and IEEE memberships
% note positions of commas and nonbreaking spaces ( ~ ) LaTeX will not break
% a structure at a ~ so this keeps an author's name from being broken across
% two lines.
% use \thanks{} to gain access to the first footnote area
% a separate \thanks must be used for each paragraph as LaTeX2e's \thanks
% was not built to handle multiple paragraphs
%

\author{Haitao~Zhang,
        Lingguo~Meng,
        Xian~Wei,~\IEEEmembership{Member,~IEEE,}
%        \cortext[mycorrespondingauthor]{Corresponding author}
        Xiaoliang~Tang,~\IEEEmembership{Member,~IEEE,} \\
         Xuan~Tang,~\IEEEmembership{Member,~IEEE,} 
        Xingping~Wang,
        Bo Jin,~\IEEEmembership{Member,~IEEE,} 
 %       Li~Fang,~\IEEEmembership{Member,~IEEE,}
        and Wei Yao,~\IEEEmembership{Member,~IEEE}
        %and~XXX~XXX,~\IEEEmembership{Life~Fellow,~IEEE}% <-this % stops a space
\thanks{ % Manuscript received . 
	This work was partially supported by CAS Pioneer Hundred Talents Program (Type C)  
	under Grant No.2017-122 and National Science Found for Young Scholars under Grant No. 61806186. 
	Xian Wei is the corresponding author, e-mail: xian.wei@fjirsm.ac.cn.
}
\thanks{H. Zhang, L. Meng and X. Wang are with the School of Software, Liaoning Technical University, Huludao 125105, China. % <-this % stops a space
L. Meng, X. Wei, X. Tang and X. Tang are with Fujian Institute of Research on the 
	Structure of Matter, Chinese Academy of Sciences, Fuzhou, 350002, China. % <-this % stops a space
%Y. L. Murphey is with Department of Electrical and Computer Engineering, 
%		University of Michigan-Dearborn, Dearborn, MI 48128, USA.
%   Y. Li is with School of Aeronautics \& Astronautics, Shanghai Jiao Tong University, 
%     Shanghai 200240, China. 
   B.Jin is with Shanghai Key Lab for Trustworthy Computing, School of Computer Science and Software Engineering, 
       East China Normal University, China.
   W. Yao is with the Department of Land Surveying and Geo-Informatics,
   The Hong Kong Polytechnic University, 181 Chatham Road South,
   Hung Hom, Kowloon, Hong Kong.
   %(e-mail:, wei.hn.yao@polyu.edu.hk).  
	} 
}

%School of Software, Liaoning Technical University, Huludao 125105, China.
% note the % following the last \IEEEmembership and also \thanks - 
% these prevent an unwanted space from occurring between the last author name
% and the end of the author line. i.e., if you had this:
% 
% \author{....lastname \thanks{...} \thanks{...} }
%                     ^------------^------------^----Do not want these spaces!
%
% a space would be appended to the last name and could cause every name on that
% line to be shifted left slightly. This is one of those "LaTeX things". For
% instance, "\textbf{A} \textbf{B}" will typeset as "A B" not "AB". To get
% "AB" then you have to do: "\textbf{A}\textbf{B}"
% \thanks is no different in this regard, so shield the last } of each \thanks
% that ends a line with a % and do not let a space in before the next \thanks.
% Spaces after \IEEEmembership other than the last one are OK (and needed) as
% you are supposed to have spaces between the names. For what it is worth,
% this is a minor point as most people would not even notice if the said evil
% space somehow managed to creep in.
% The paper headers
\markboth{1D-Convolutional Capsule Network for Hyperspectral Image Classification}%
{Zhang and Meng \MakeLowercase{\textit{et al.}}: \textit{1D-ConvCapsNet} for HSI Classification}
% The only time the second header will appear is for the odd numbered pages
% after the title page when using the twoside option.
% 
% *** Note that you probably will NOT want to include the author's ***
% *** name in the headers of peer review papers.                   ***
% You can use \ifCLASSOPTIONpeerreview for conditional compilation here if
% you desire.

% If you want to put a publisher's ID mark on the page you can do it like
% this:
%\IEEEpubid{0000--0000/00\$00.00~\copyright~2015 IEEE}
% Remember, if you use this you must call \IEEEpubidadjcol in the second
% column for its text to clear the IEEEpubid mark.

% use for special paper notices
%\IEEEspecialpapernotice{(Invited Paper)}

% make the title area
\maketitle

% As a general rule, do not put math, special symbols or citations
% in the abstract or keywordsss
\begin{abstract}
Recently, convolutional neural networks (CNNs) have achieved excellent performances in many computer vision tasks. 
Specifically, for hyperspectral images (HSIs) classification, CNNs often require very complex structure due to the high dimension of HSIs. 
The complex structure of CNNs results in prohibitive training efforts. 
%
%With the structure deepening, conventional CNNs suffer from gradient disappearance (GD), spatial information damage and prohibitive training efforts. 
%
%For hyperspectral images (HSIs) classification task, convolutional neural networks (CNNs) often require very complex structure due to the high dimension of HSIs, resulting in prohibitive training efforts. 
%
Moreover, the common situation in HSIs classification task is the lack of labeled samples, which results in accuracy deterioration of CNNs. 
In this work, we develop an easy-to-implement capsule network to alleviate the aforementioned problems, i.e., \textit{1D-convolution capsule network (\textit{1D-ConvCapsNet})}. 
%
%in order to solve problems above, a novel capsule-based network is proposed, namely, \textit{1D-convolution capsule network (\textit{1D-ConvCapsNet})}, with expect to increase the classification accuracy and reduce the training efforts. 
%
Firstly, \textit{1D-ConvCapsNet} separately extracts spatial and spectral information on spatial and spectral domains, which is more lightweight than 3D-convolution due to fewer parameters. 
%we propose a lightweight strategy to extract spectral-spatial information in a separable way, which allows features extraction to be performed on spatial and spectral domains respectively. 
%
Secondly, \textit{1D-ConvCapsNet} utilizes the capsule-wise \textit{constraint window} method to reduce parameter amount and computational complexity of conventional capsule network. 
%
%Secondly, inspired by 1D-convolution, we utilize the \textit{constraint windows} method to reduce the number of parameters and the complexity of conventional capsule network. 
%
%This can be viewed as the extension of 1D-convolution on capsule neurons. 
%
Finally, \textit{1D-ConvCapsNet} obtains accurate predictions with respect to input samples via dynamic routing. 
The effectiveness of the \textit{1D-ConvCapsNet} is verified by three representative HSI datasets. 
%
%several experiments are conducted to verify the effectiveness of the proposed \textit{1D-ConvCapsNet}. 
%
%In the experiment, several well-known methods are compared on three representative HSI datasets. 
%
Experimental results demonstrate that \textit{1D-ConvCapsNet} is superior to state-of-the-art methods in both the accuracy and training effort. 
\end{abstract}

% Note that keywords are not normally used for peerreview papers.

% For peer review papers, you can put extra information on the cover
% page as needed:
% \ifCLASSOPTIONpeerreview
% \begin{center} \bfseries EDICS Category: 3-BBND \end{center}
% \fi
%
% For peerreview papers, this IEEEtran command inserts a page break and
% creates the second title. It will be ignored for other modes.
\IEEEpeerreviewmaketitle

\section{Introduction}
\label{sec:01}
% The very first letter is a 2 line initial drop letter followed
% by the rest of the first word in caps.
% 
% form to use if the first word consists of a single letter:
% \IEEEPARstart{A}{demo} file is ....
% 
% form to use if you need the single drop letter followed by
% normal text (unknown if ever used by the IEEE):
% \IEEEPARstart{A}{}demo file is ....
% 
% Some journals put the first two words in caps:
% \IEEEPARstart{T}{his demo} file is ....
% 
% Here we have the typical use of a "T" for an initial drop letter
% and "HIS" in caps to complete the first word.
\IEEEPARstart{U}{nlike} the colorful image only responding to visible light, HSI contains hundreds of spectral channels (SCs), each of which is an image of the target in a very narrow segment of the electromagnetic spectrum. 
HSI describes object's surface with abundant spectral-spatial information \cite{IEEEhowto:1_1_1,IEEEhowto:1_1_2,IEEEhowto:1_1_3,IEEEhowto:1_1_4}. 
Thus, HSI has become increasingly popular in remote sensing fields, such as ecological science, precision agriculture and mineral exploration, etc. \cite{IEEEhowto:1_1_2,IEEEhowto:1_1_5,IEEEhowto:1_1_6,IEEEhowto:1_1_7,IEEEhowto:1_1_8,IEEEhowto:1_1_9}. 
The classification of HSIs is the key to realize the applications above.

In spite of lots of research efforts have been attracted on the aforementioned fields, certain essential characteristics of HSIs make it very challenging for the classification task. 
Basically, the main challenging characteristics can be summarized as follows. 
\begin{itemize}
	\item [1)] 
	HSI usually consists of several hundreds of SCs, and every pixel has high dimensionality. Hence, HSI has high structural complexity due to the so-called curse of dimensionality (Hughes phenomenon\cite{IEEEhowto:1_1_1}).    
	\item [2)]
	The quality of HSI is affected by multi-factors, such as weather and illumination. These factors often generate the confusion of HSI, which result in the phenomenon that the same spectrum expresses different objects, or same objects have different spectrum. 
	\item [3)]
	The number of labeled training samples are often limited because labeling data is expensive and time consuming. Consequently, finite training samples cannot generalize the whole of ground-truth, resulting in degradation of common classifiers.
\end{itemize}

Researchers are motivated by these challenges to develop more effective methods. 
Previously, support vector machine (SVM), sparse representation classifier, $k$-means clustering were employed to classify the HSIs by only using multi-bands spectral information \cite{IEEEhowto:1_3_1,IEEEhowto:1_3_2,IEEEhowto:1_3_3}.  
In order to capture more useful information,
recently, deep learning (DL)-based \cite{IEEEhowto:1_3_4} methods exhibited evident advantage on HSIs classification because of their capacity of feature extraction from low-level to high-level. 
The method of stacked autoencoder (SAE) for HSIs data was proposed in \cite{IEEEhowto:1_3_5}. 
Another deep learning method was also proposed using deep belief network (DBN) in \cite{IEEEhowto:1_3_6}. 
These methods extract the global discriminative feature but ignoring the spatial correlation, suffering from the highly computational complexity. 
Deep feature extraction and classification for HSIs using CNN was introduced by \cite{IEEEhowto:1_3_7}, which achieved the state-of-the-art performance due to its ability to extract local spatial relationship.

Generally, CNNs extract features through flexible combination of convolution and pooling layers. 
This is a simplified process of the comprehension of the brain to visual stimulation from retina to cortex. 
The flexibility of CNN structure make it suitable for computer vision applications, such as images classification and objects detection \cite{IEEEhowto:1_4_1,IEEEhowto:1_4_2}. 
In HSIs classification, CNNs also hold outstanding performance because 3D-convolution \cite{IEEEhowto:1_4_3} can extract spectral-spatial information effectively \cite{IEEEhowto:1_3_7,IEEEhowto:1_4_4,IEEEhowto:1_4_5,IEEEhowto:1_4_6}. 
However, this outstanding performance relies heavily on complex network structure due to the characteristics of HSI data. 
Conventional CNNs suffer from gradient disappearance (GD), spatial information damage and prohibitive training efforts.
On the one hand, the training process is extremely deferred due to the GD. 
On the other hand, the spatial information is unavoidably damaged due to down-sampling. 
Therefore, CNN require massive data and additional epoch during the training phase. 
Obviously, under this straightforward CNN mechanism, the training efforts will be undoubtedly increased.

For the issues mentioned above, a novel type of neural network was proposed in 2017, namely, CapsNet \cite{IEEEhowto:1_5_1}, which only contains three layers. 
It replaces scalar neuron of CNN with vector neuron, and replaces pooling of CNN with dynamic routing for the representation  of the part-whole relationship of data. 
In this way, the CapsNet has gained considerable generalization capabilities. 
CapsNet has achieved the state-of-the-art performance on MNIST and also achieved the outstanding performance on CIFAR-10. 
The latest studies reveal its potential on image segmentation, 3D vision and object detection \cite{IEEEhowto:1_5_2,IEEEhowto:1_5_3,IEEEhowto:1_5_4}. 
The CapsNet provides a new way of deep learning for researchers, but it suffers from memory burden and low training speed due to parameters redundancy. 
In this paper, we develop lightweight methods for the capsule network and build a novel architecture for HSIs classification. 
Compared to conventional methods, our method is highly competitive in terms of accuracy and training efforts. 
The main contributions of this paper are summarized as follows. 
\begin{itemize}
	\item [1)]
	It proposes a separated spatial and spectral information extraction method, which extracts features on spatial and spectral domains, respectively. 
	\item [2)]
	It proposes a \textit{constraint window} method to reduce the complexity of capsule network while holding accuracy. 
	\item [3)]
	It builds a accurate and efficient capsule network by using our proposed methods, called \textit{1D-ConvCapsNet}. 
\end{itemize}

The rest of paper is organized as follows, Section \ref{sec:02} introduces related works. 
Section \ref{sec:03} describes classification strategy of our method. 
Section \ref{sec:04} gives architecture and detailed implementation. 
Section \ref{sec:05} presents the results of experiments. 
Finally, conclusions are given in Section \ref{sec:06}.

\section{Related Works}
\label{sec:02}
HSIs classification is the fundamental task in remote sensing application. 
Although HSIs classification has been studied extensively by researchers from multiple perspectives in the past, inherent characteristics of HSIs make it a challenging task. 
Previously, the SVM is widely used because of its effectiveness and robustness. 
It projects samples into high-dimensional feature space using kernel-based method to make samples linearly separable. 
Specifically, \cite{IEEEhowto:1_3_1} employed kernel trick that promoted the separation of samples in a high-dimensional feature space via a nonlinear transformation of a kernel function. 
With the improvement of spectral resolution, conventional approaches suffer from Hughes phenomenon \cite{IEEEhowto:1_1_1}. 
In order to deal with this phenomenon, a series of dimensionality reduction (DR) and band selection (BS) approaches were developed by researchers.
\cite{IEEEhowto:2_1_1,IEEEhowto:2_1_2} proposed image low-dimensional representation learning method for reconstructible DR under unsupervised conditions. 
\cite{IEEEhowto:2_1_3} presented the locality adaptive discriminant analysis method for HSIs classification. 
\cite{IEEEhowto:2_1_4} proposed the salient BS method via manifold ranking. 
Recently, DL-based methods have been widely adopted  because they are capable of automatically learning features. 
Compared to conventional methods, DL-based methods have hierarchical structure, which generate high-level features from low-level features via forward propagation. 
Typically, several DL-based HSIs classification methods were proposed. 
\cite{IEEEhowto:1_3_5} proposed a spectral-spatial joint information method by combining SAE with PCA. 
\cite{IEEEhowto:2_1_6} proposed a unsupervised features extraction method which could also be used to extract spectral-spatial information. 
However, for the above methods, the input data has to be reshaped to vector, which result in the loss of spatial correlation. 
In order to exploit spatial correlation effectively, CNN has become one of the most important tool in remote sensing because CNN can naturally extract multi-dimensional features. 
Especially, 3D-convolution \cite{IEEEhowto:1_4_3} can extract features simultaneously in spectral and spatial domains of HSI. 
Hence, several CNN-based approaches have been proposed. 
\cite{IEEEhowto:2_1_8} designed a dual branches end-to-end network with skip architecture to learn spectral and spatial features, respectively. 
\cite{IEEEhowto:2_1_9} proposed a joint features extraction method with two branches, which are devoted to features from the spectral domain and the spatial domain. 
\cite{IEEEhowto:2_1_10} proposed semi-supervised network with skip connection between the encoder and the decoder in order to solve the problem of limited labeled samples. 
\cite{IEEEhowto:2_1_11} proposed a HSIs classification method with Markov random fields and CNN from the perspective of unified Bayesian framework. 
\cite{IEEEhowto:1_3_7} proposed a series of regularized deep feature extraction methods using several convolution and pooling layers, which achieved the state-of-the-art performance. 
%

%
%Needless to say,
CNN has become a powerful tool for HSIs classification task. 
However, compared to ordinary images classification task, CNN-based methods have more complex structures due to the complexity of HSI data, which require exorbitant efforts during the training phase. 
Firstly, with the structure deepening, the gradient is gradually lost during the propagation process, resulting in the slow convergence rate. 
\cite{IEEEhowto:2_2_1} introduced the rectified linear unit ($ReLU$) as the activation function in order to alleviate the GD. 
Based on this, \cite{IEEEhowto:1_4_1} designed AlexNet with $ReLU$ activation function and won the annual ImageNet competition. 
Secondly, CNNs often use pooling to control scale of networks, which inevitably damage the spatial information due to down sampling. 
To deal with this, \cite{IEEEhowto:2_2_2} proposed a global pooling method, called dynamic $k$-max pooling, which keeps top-$k$ values during the pooling operation. 
\cite{IEEEhowto:2_2_3} presented the architecture with overlap pooling for HSIs classification by using different combinations of max pooling and mean pooling. 
Thirdly, CNN-based methods are unable to detect pose information of the objects because convolution filters can only represent the activity related to features. 
It means that CNNs are invariant for spatial transformation of objects. 
Hence, CNNs are incapable of modeling relative relationship between objects. 
Some data augmentation methods have been adopted in order to make CNNs more robust with respect to spatial transformation and prevent overfitting under the limited training samples\cite{IEEEhowto:1_3_7,IEEEhowto:2_2_4}.

In order to overcome the shortcomings of traditional CNN, \cite{IEEEhowto:2_3_1} proposed the conception of capsule, which can encode instantiation parameters of entity (an object or object part).
Capsule is a collection of neurons, which describes the pose information and existence probability of an entity. 
Hence, the capsule carries more information about properties of the entity than the conventional scalar neuron of CNN. 
In order to effectively use information stored in the capsule neurons, \cite{IEEEhowto:1_5_1} proposed dynamic routing between capsules and designed a novel network, called CapsNet. 
Therein, a capsule neuron is organized into a vector, whose length and orientation respectively represent existence probability and properties of an entity. 
Dynamic routing is used for communication between capsules by adjusting 
the coupling coefficient between predictive vector and high-layer capsule. 
The coupling indicates that entity in the image should be paid attention to rather than directly encoding it. 
Therefore, the capsule-based network is more expressive and explanatory than the conventional CNN. 
Specifically, by taking advantages of capsules, capsule-based networks exhibit high precision, fast convergence, strong noise immunity and generalization.

However, CapsNet has massive training parameters, which result in high storage pressure and slow training speed. 
Parameter redundancy makes CapsNet difficult to directly work on large images. 
In this paper, \textit{1D-ConvCapsNet} was proposed, which is an easy-to-implement method for HSIs classification task. 
The details of \textit{1D-ConvCapsNet} will be describe in Section \ref{sec:04}.

\section{Classification Strategy}
\label{sec:03}
\subsection{CNN-based model}
\label{sec:031}
The CNN-based model is popular in HSIs analysis and processing because the convolution can easily extract features on multi-dimension. 
As shown in Fig. \ref{fig:struCNN}, a classic CNN model mainly contains two modules, features extraction module (FEM) and classifier module (CM). 
Generally, FEM uses combination of the convolution and the pooling layers to extract high-level features, and CM uses several full connection layers as a classifier. 
In CNN, the convolution layer is the most important component, which is related to two aspects. 
One aspect is the statistical properties of images\cite{IEEEhowto:3_1_1}, that means features learned at a region can be applied to others. 
This fact allows convolution filters to detect the same features at all position of images. 
Another aspect is the finding of neuroscience\cite{IEEEhowto:3_1_2}, which reveals that cells within receptive fields of vision system are sensitive to visual stimulus 
and have strong responses to interested features. 
Additionally, visual cells are mainly composed of two types, s-cells and c-cells. 
The s-cells have an intensively response to their preferences, which functionally correspond to the convolution layer. 
The c-cells are able to concentrate multiple s-cells to achieve large receptive fields and resist distortion, which functionally correspond to the pooling layer. 
Therefore, the classic CNN model can be regarded as the oversimplified simulations of the visual system. 

Concretely, convolution filters implement the aforementioned ideas in a manner that locally connect and share parameters. 
Taking 3D-convolution as an example, the convolution operation of one filter at position $(x,y,z)$ in layer $l$ can be defined as follows:
\begin{figure}
	\includegraphics{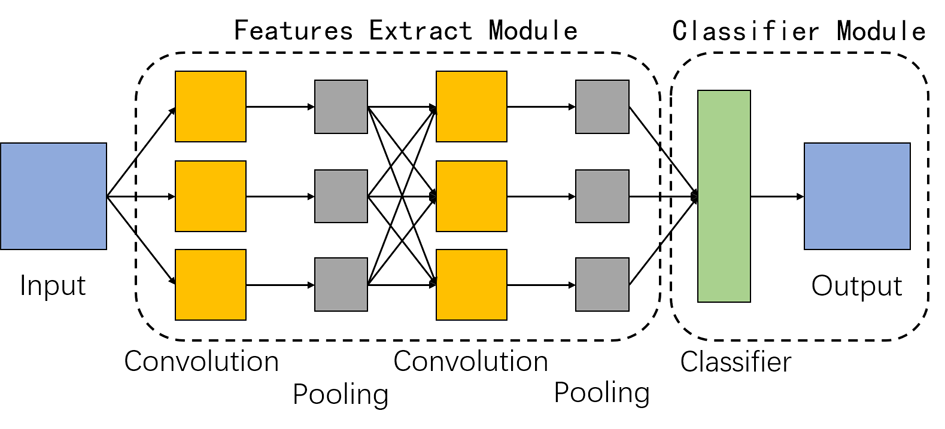}
	\centering
	\caption{The classic structure of CNN. It consists of two modules: Feature extraction module and classifier module. }
	\label{fig:struCNN} 
\end{figure}
\begin{equation}
\label{eq:3DConv}
\begin{split}
v_{x,y,z}^{(l)}&=\sigma(\sum_{m}\sum_{i=0}^{\alpha-1}\sum_{j=0}^{\beta-1}\sum_{k=0}^{\gamma-1}w_{i,j,k}^{(l),m}v_{x+i,y+j,z+k}^{(l-1),m}+b^{(l)})\\
\end{split}
\end{equation}
where $v_{x,y,z}^{(l)}$ is the activity of neuron at position $(x,y,z)$ in a feature map of layer $l$. 
The $m$ is index of the feature maps generated by the previous layer. 
Constant $\alpha$, $\beta$ and $\gamma$ represent the spatial size of convolution filter.
In HSIs classification task, $\alpha$ and $\beta$ correspond to the spatial domain, and $\gamma$ corresponds to the spectral domain. 
The $w_{i,j,k}^{(l),m}$ is weight parameter at $(i,j,k)$ of convolution filter corresponding to the $m$-th feature map in layer $l$. 
The $b^{(l)}$ is the bias parameter of convolution filter in layer $l$. 
All $w_{i,j,k}^{(l),m}$ and $b^{(l)}$ are trained by back propagation (BP) algorithm.

Function $\sigma(\cdot)$ is defined as the nonlinear activation function. 
It introduces the nonlinearity into neural network (NN) for enhancing performance. 
Function $ReLU$ is widely used because of its advantages of simplicity, rapidity and avoiding GD. 
It is given by the following equation:
\begin{equation}
\label{eq:ReLU}
\begin{split}
\sigma(x)={\rm{max}}(0, x).
\end{split}
\end{equation}

The pooling layer is located behind the convolution layer, which provides a larger receptive field and a degree of transform invariance through down sampling. 
The max-pooling operation is defined as follows:
\begin{equation}
\label{eq:maxP}
\begin{split}
o_{x,y,z}={\rm{max}}(\textbf{R}_{x,y,z}\textbf{V})
\end{split}
\end{equation}
where the $\textbf{V}$ is defined as the input feature maps generated by previous convolution layer, 
and the $\textbf{R}_{x,y,z}$ is defined as a operator that extracts patch from $\textbf{V}$ at position $(x,y,z)$. 
The $o_{x,y,z}$ is the maximum in the patch.

\subsection{Capsule-based model}
\label{sec:032}
Different from CNN-based model, the basic unit of the capsule-based network is capsule neuron, which consists of several scalars. 
In CapsNet, capsule neuron first describes an entity in the form of a vector, which holds the existence probability and properties of an entity. 
Specifically, properties are expressed as instantiation parameters, i.e., pose (position, size, orientation), deformation, texture, etc. 
Then, a viewpoint-invariant representation can be obtained by multiplying viewpoint matrix and capsule. 
This idea stems from computer graphics, which can be understood as the inverse rendering process. 
The rendering process is to give an abstract representation and instantiation parameters of the entity, and then get the image by using the render function. 
Capsule is in an opposite way, it acquires approximately abstract representation of entity via viewpoint matrix and instantiation parameters. 
This process has the viewpoint invariance, meaning that whether the direction of observation changes, the abstract representation of entity can be obtained by using the same viewpoint matrix. 

\begin{figure}
	\includegraphics{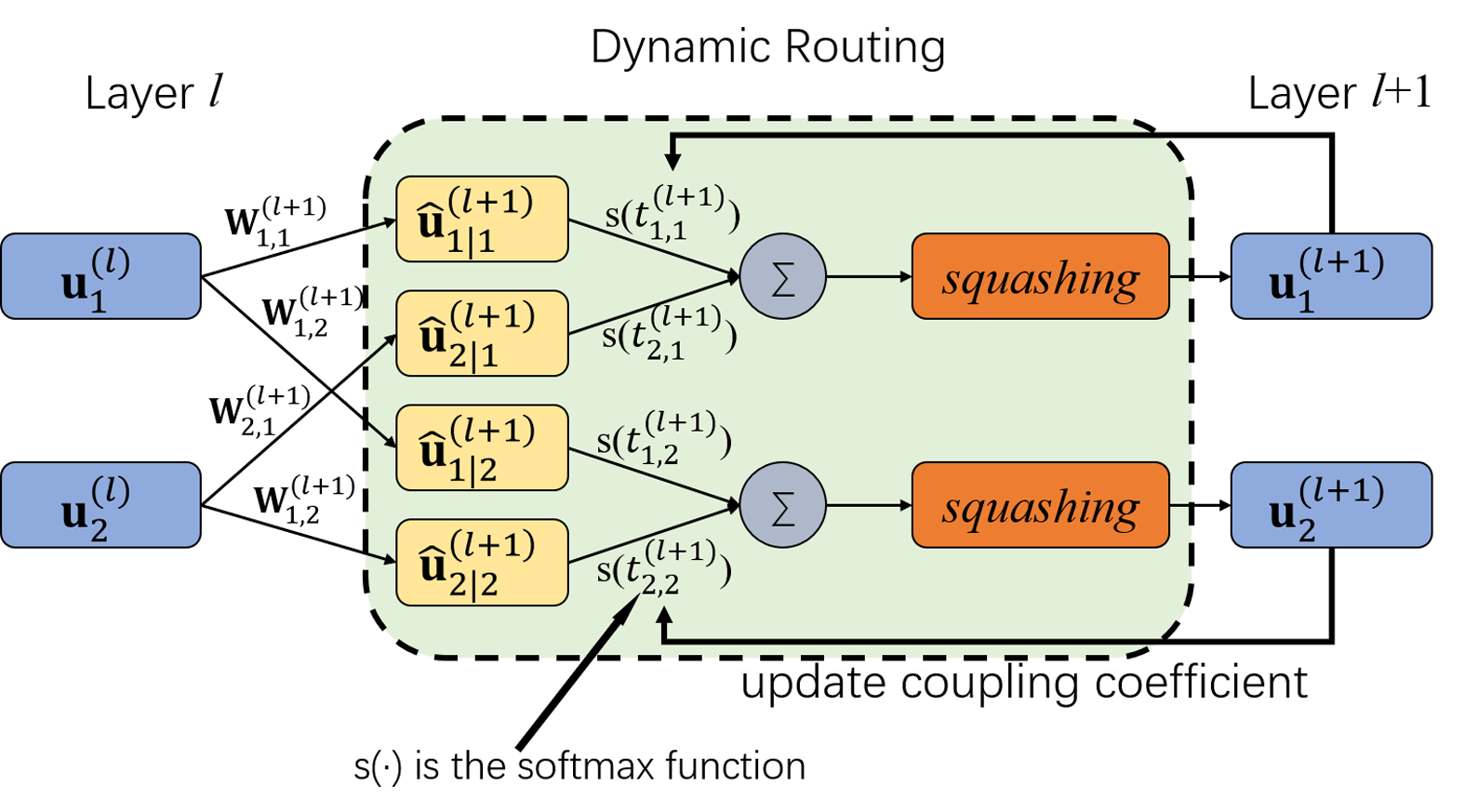}
	\centering
	\caption{The execution of dynamic routing. It iteratively updates the coupling coefficient by accumulating $t_{i,j}^{(l+1)}$, where $\textbf{u}_{i}^{(l)}$, $\hat{\textbf{u}}_{j\mid i}^{(l+1)}$, $\textbf{u}_{j}^{(l+1)}$ and $t_{i,j}^{(l+1)}$ represent child capsule, prediction capsule, parent capsule and log prior, respectively.}
	\label{fig:exeDR} 
\end{figure}

Dynamic routing \cite{IEEEhowto:1_5_1} is the most important part of the capsule-based network. 
It determines the coupling coefficient by measuring the agreement between capsules, allowing them to be dynamically connected. 
This mechanism make child capsules more inclined to send messages to the parent capsule with large coupling, and child capsules also receive the feedback from parent that indicates which entity in the image should be paid attention to. 
Intuitively, the execution of dynamic routing is shown in Fig. \ref{fig:exeDR},
where child capsule is denoted by $\textbf{u}_{i}^{(l)}\in\mathbb{R}^{d^{(l)}}$ and parent capsule is denoted by $\textbf{u}_{j}^{(l+1)}\in\mathbb{R}^{d^{(l+1)}}$. 
The viewpoint-invariant representation is denoted by $\hat{\textbf{u}}_{j\mid i}^{(l+1)}\in\mathbb{R}^{d^{(l+1)}}$, which is also called the prediction vector. 
Firstly, prediction vector $\hat{\textbf{u}}_{j\mid i}^{(l+1)}$ is obtained by 
multiplying $\textbf{W}_{i,j}^{(l+1)}\in\mathbb{R}^{d^{(l+1)}\times d^{(l)}}$ by $\textbf{u}_{i}^{(l)}$, and log prior $t_{i,j}^{(l+1)}$ is initialized to zero. 
Secondly, $\textbf{u}_{j}^{(l+1)}$ is equal to weighted sum of all $\hat{\textbf{u}}_{j\mid i}^{(l+1)}$ through the $squashing$ nonlinear activation function. 
Thirdly, the log prior $t_{i,j}^{(l+1)}$ is updated by the accumulation of the scalar product between $\hat{\textbf{u}}_{j\mid i}^{(l+1)}$ and $\textbf{u}_{j}^{(l+1)}$. 
By iterating the second and the third steps, the coupling coefficient can be allocated to achieve dynamic connection between capsules. 
Similar to CNN, the viewpoint matrix $\textbf{W}_{i,j}^{(l+1)}$ can be learned by BP algorithm.

\begin{figure}
	\includegraphics{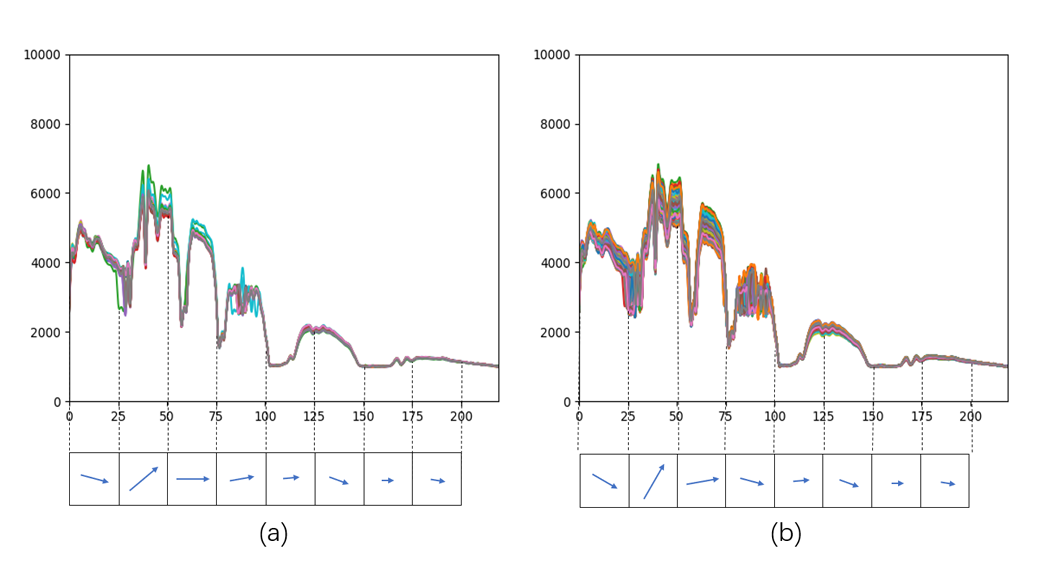}
	\centering
	\caption{Entity differences between two classes of ground-truth. Suppose the capsule represents an entity with every 25 spectral channels. The lengths of vector are similar, but their directions are different due to different instantiation parameters.}
	\label{fig:differGT} 
\end{figure}

In the HSIs classification task, this mechanism also can be used, but the key is how to express entity with capsule. 
In this work, we define a part of the SCs as an entity. 
Capsule can encode the existence probability and instantiation parameters of the entity using the length and direction of the vector, respectively. 
As shown in Fig. \ref{fig:differGT}, each type of ground-truth has its own characteristics in the representation of capsules.
Hence, these characteristics can be interpreted by dynamic routing during program execution. 

\section{The proposed method}
\label{sec:04}
\begin{figure*}
	\centering
	\includegraphics{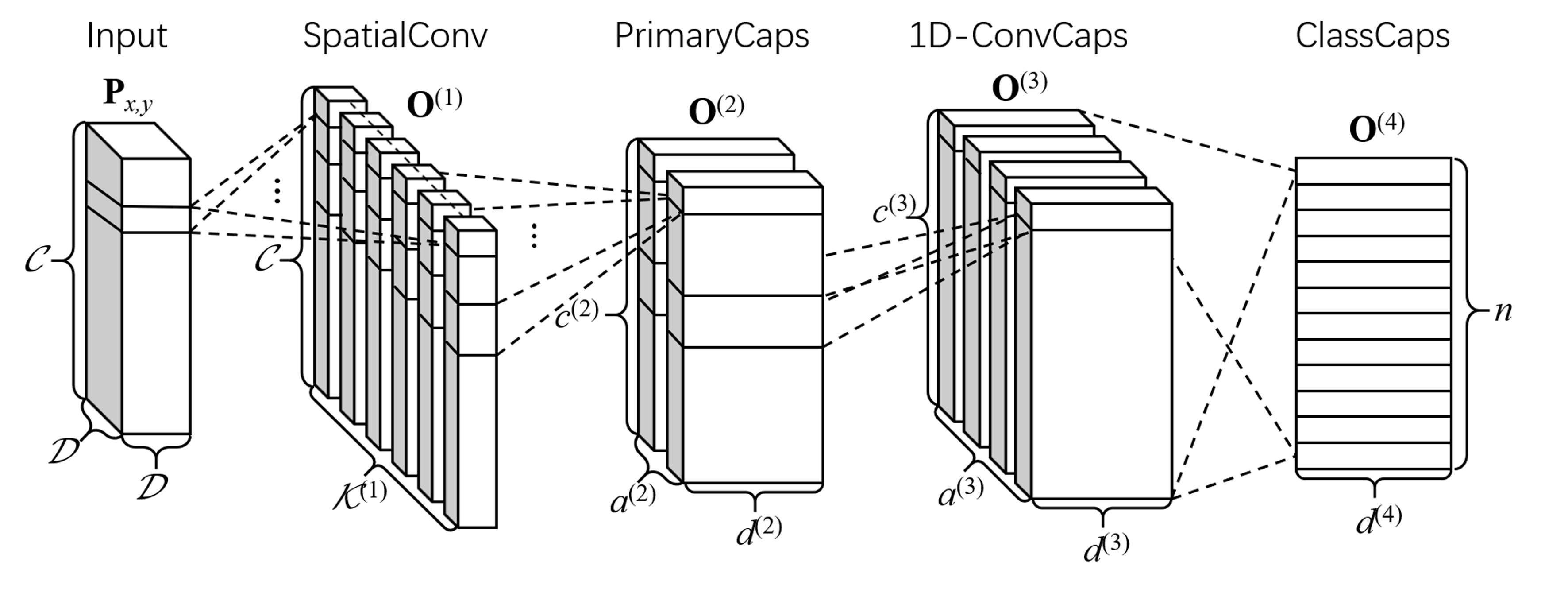}
	\caption{The architecture of our proposed method. It consist of four layers, SpatialConv layer is used to extract spatial information; PrimaryCaps layer is used to extract spectral information and form capsule units; 1D-ConvCaps layer is used to reduce parameters; ClassCaps layer is used to represent part-whole relationship.}
	\label{fig:archi1DCCN} 
\end{figure*}

In this section, we discuss architecture and implementation details of the proposed \textit{1D-ConvCapsNet}. 
Firstly, \textit{1D-ConvCapsNet} extracts spectral-spatial information by using our proposed method, which extracts information on spatial and spectral domains respectively to form capsule units. 
However, in conventional CNN-based models, 3D-convolution is used to extract spectral-spatial information, which needs more expensive efforts in terms of computation and storage than proposed method. 
Secondly, \textit{constraint window} is used to reduce the number of parameters, which are inspired by local strategy of convolution. 
The \textit{constraint window} limits the generation of parent vector in the local regions, which combines simple entities into complex entity and provides greater receptive field.  
Finally, \textit{1D-ConvCapsNet} uses the dynamic routing to combine complex entities to the whole. 
\textit{1D-ConvCapsNet} consists of four layers, which structure shown as Fig. \ref{fig:archi1DCCN}. 

The HSI can be viewed as a data cube ${\textbf{X}}\in\mathbb{R}^{\mathcal{H}\times \mathcal{W}\times \mathcal{C}}$, which consist of $\mathcal{H}\times \mathcal{W}$ pixels. 
Each pixel ${\textbf{x}}_{x,y}\in\mathbb{R}^{\mathcal{C}}$ belongs to a class of ground truth, providing abundant spectral information. 
Moreover, each pixel and its neighborhoods often belong to the same class, providing spatial information.  
Hence, the performance of classifier is improved by considering spectral and spatial information. 
Based on this fact, a patch $\textbf{P}_{x,y}\in \mathbb{R}^{\mathcal{D}\times \mathcal{D}\times \mathcal{C}}$ of HSI is picked, which sample is located at the center point $(x,y)$. 
In other words, the picked patch is the $\mathcal{D}\times \mathcal{D}\times \mathcal{C}$ data block. 
\textit{1D-ConvCapsNet} extracts spectral-spatial information from input data blocks.

\subsection{SpatialConv Layer}
\label{sec:041}
This layer is the first layer of network, represented by $\mathcal{L}^{(1)}$, the input of which is patch $\textbf{P}_{x,y}$.  
It extracts the spatial information by using the same 2D-convolution filter with size $f^{(1)}=\mathcal{D}\times \mathcal{D}$ on each SC of HSI. 
In this layer, only the spatial information in respective SC is extracted. 
Therefore, the SpatialConv layer applies $\mathcal{K}^{(1)}$ 2D-convolution filters on each SC of the $\textbf{P}_{x,y}$ to obtain $\mathcal{K}^{(1)}$ feature maps, each of which has size $1\times 1\times \mathcal{C}$. 
These feature maps are output of this layer, denoted by ${\textbf{O}}^{(1)}\in \mathbb{R}^{\mathcal{C}\times \mathcal{K}^{(1)}}$.

\subsection{PrimaryCaps Layer}
\label{sec:042}
This layer is the first of capsule layers, represented by $\mathcal{L}^{(2)}$, the input of which is ${\textbf{O}}^{(1)}$. 
Its goal is to extract spectral information from ${\textbf{O}}^{(1)}$ and form capsule units. 
Firstly, the PrimaryCaps layer applies $\mathcal{K}^{(2)}$ 1D-convolution filters with size $f^{(2)}$ on ${\textbf{O}}^{(1)}$ to obtain $\mathcal{K}^{(2)}$ feature maps. 
Secondly, these feature maps are stacked into a tensor with size $c^{(2)}\times a^{(2)}\times d^{(2)}$, where $a^{(2)}\times d^{(2)}=\mathcal{K}^{(2)}$.
This tensor is the output of PrimaryCaps layer, denoted by ${\textbf{O}}^{(2)}\in \mathbb{R}^{c^{(2)}\times a^{(2)}\times d^{(2)}}$. 
The $c^{(2)}$ is the number of capsules in each capsule array, $a^{(2)}$ is the number of capsule arrays, and $d^{(2)}$ is the dimension of every capsule.

Compared to the 3D-convolution, our proposed spectral-spatial information extraction method is more efficient because this separated method can reduce parameter redundancy.
Compare to the scalar neuron of CNN-based model, capsule is more expressive because additional information is stored. 
Therefore, the capsule network has great advantages in noise immunity, convergence speed and generalization ability.
From this layer, the basic unit of network is capsule neuron, which connects to 1D-ConvCaps layer through \textit{constraint window}.

\subsection{1D-ConvCaps Layer}
\label{sec:043}
This layer is used to reduce the number of parameters, represented by $\mathcal{L}^{(3)}$. 
In CapsNet, full connection is used directly between adjacent capsule layers, resulting in parameters redundancy.  
Hence, CapsNet has high storage pressure and time cost due to numerous parameters. 
Herein, we propose the \textit{constraint window} method for capsule network, which utilizes a local strategy to reduce parameters redundancy of the network. 
\textit{Constraint window} realizes the goal through local connection and sharing parameters.

This layer constructs $\mathcal{K}^{(3)}$ \textit{constraint windows} on input ${\textbf{O}}^{(2)}$ for generating $\mathcal{K}^{(3)}$ output capsule arrays, 
denoted by $\textbf{O}^{(3)}\in \mathbb{R}^{c^{(3)}\times a^{(3)}\times d^{(3)}}$, where $a^{(3)}=\mathcal{K}^{(3)}$. 
The size of each window is $f^{(3)}\times a^{(2)}\times c^{(2)}$, where $f^{(3)}$ is the artificially specified size.  
The $q$-th window has a viewpoint tensor $\textbf{T}_{q}^{(3)}\in \mathbb{R}^{d^{(3)}\times f^{(3)}\times a^{(2)}\times d^{(2)}}$, which does not depend on a fixed spatial location and shares parameters with other child capsules.
For the parent capsule $\textbf{u}_{q,k}^{(3)}\in {\textbf{O}}^{(3)}$ at the $k$-th position of the $q$-th capsule array, it is equal to the product of $\textbf{T}_{q}^{(3)}$ and $\textbf{C}_{k}^{(2)}\in \mathbb{R}^{f^{(3)}\times a^{(2)}\times d^{(2)}}$. 
Formally, $\textbf{u}_{q,k}^{(3)}$ expressed by following equation:
\begin{equation}
\label{eq:parV1}
\begin{split}
\textbf{u}_{q,k}^{(3)}=\textbf{T}_{q}^{(3)}\times \textbf{C}_{k}^{(2)}+\textbf{b}_{q}^{(3)}
\end{split}
\end{equation}
where $\textbf{C}_{k}^{(2)}$ is child capsules of $\textbf{O}^{(2)}$ covered by \textit{constraint window} at position $k$ and $\textbf{b}_{q}^{(3)}$ is learned bias for parent capsule. 
Moreover, $\textbf{T}_{q}^{(3)}$ can be decomposed into $f^{(3)}\times a^{(2)}$ viewpoint matrices $\textbf{W}_{q,i,j}^{(3)}\in \mathbb{R}^{d^{(3)}\times d^{(2)}}$, where $i$ and $j$ are indices of $a^{(2)}$ and $f^{(3)}$ respectively. 
Hence, $\textbf{u}_{q,k}^{(3)}$ is the sum of product of viewpoint matrix and corresponding child capsule. It is formulated as follows: 
\begin{equation}
\label{eq:parV2}
\begin{split}
\textbf{u}_{q,k}^{(3)}=\sum_{i=1}^{a^{(2)}}\sum_{j=1}^{f^{(3)}} {\textbf{W}_{q,i,j}^{(3)}\times \textbf{c}_{i,(k-1)s+j}^{(2)}}+\textbf{b}_{q}^{(3)}
\end{split}
\end{equation}
where $s$ is the stride of \textit{constraint window} move to next position. 
Similar to conventional convolution, the viewpoint matrix can be learned by BP algorithm.

\subsection{ClassCaps Layer}
\label{sec:044}
This layer is a dense capsule layer, which represents the part-whole relationship of capsules, denoted by $\mathcal{L}^{(4)}$. 
Its output ${\textbf{O}}^{(4)}\in \mathbb{R}^{n\times d^{(4)}}$ is obtained by using dynamic routing on ${\textbf{O}}^{(3)}$. 
Every capsule $\textbf{u}_{k}^{(4)}\in \textbf{O}^{(4)}$ represents a class of ground truth in HSI, where the length of vector is the probability of sample belonging to the corresponding class. 
Hence, the capsule $\textbf{u}_{k}^{(4)}$ is also referred as the activation capsule. 
In order to obtain the activation capsule from $\mathcal{L}^{(3)}$, firstly, dynamic routing needs to calculate the weighted sum of prediction vectors by following equation:
\begin{equation}
\label{eq:actV}
\begin{split}
\textbf{s}_{k}^{(4)}=\sum_{i=1}^{a^{(3)}}\sum_{j=1}^{c^{(3)}} l_{i,j,k}^{(4)}\times \hat{\textbf{u}}_{k\mid i,j}^{(4)}.
\end{split}
\end{equation}
Therein, prediction vector $\hat{\textbf{u}}_{k\mid i,j}^{(4)}$ is equal to product of the child capsule $\textbf{u}_{i,j}^{(3)}$ and the corresponding viewpoint matrix $\textbf{W}_{i,j,k}^{(4)}$, and $l_{i,j,k}^{(4)}$ is the coupling coefficient between $\hat{\textbf{u}}_{k\mid i,j}^{(4)}$ and $\textbf{u}_{k}^{(4)}$.
Formally, $l_{i,j,k}^{(4)}$ is computed by a function of $routing$ $softmax$ as follows:
\begin{equation}
l_{i,j,k}^{(4)}=\frac{\textup{exp}(t_{i,j,k}^{(4)})}{\sum_{g=1}^{n} \textup{exp}(t_{i,j,g}^{(4)})}
\end{equation}
where $t_{i,j,k}^{(4)}$ is the log prior of $\hat{\textbf{u}}_{k\mid i,j}^{(4)}$ coupling to $\textbf{u}_{k}^{(4)}$ and initial value is zero. 
Secondly, capsule $\textbf{u}_{k}^{(4)}$ is computed by a nonlinear activation function, called $squashing$:
\begin{equation}
\textbf{u}_{k}^{(4)}=\frac{||\textbf{s}_{k}^{(4)}||^{2}}{1+||\textbf{s}_{k}^{(4)}||^{2}}\frac{\textbf{s}_{k}^{(4)}}{||\textbf{s}_{k}^{(4)}||}.
\end{equation}
Finally, the agreement between $\hat{\textbf{u}}_{k\mid i,j}^{(4)}$ and $\textbf{u}_{k}^{(4)}$ is measured by the simple scalar product $\langle \hat{\textbf{u}}_{k\mid i,j}^{(4)},\textbf{u}_{k}^{(4)} \rangle$. 
A large scalar product means the increasing of coupling coefficient between capsules. 
By iterating the above process and accumulate $t_{i,j,k}^{(4)}$, the coupling coefficient $l_{i,j,k}^{(4)}$ can be obtained rapidly. 
This not only transmits information between capsules, but also connects parts to the whole by assigning coupling coefficients.

\subsection{Loss Function}
\label{sec:045}
In training phase, a patch $\textbf{P}_{x,y}$ is input into the network and $n$ activity vectors are obtained by forward propagation. 
Before the back propagation, the network needs to measure the gap between $\textbf{O}^{(4)}$ and label via loss function. 
Herein, we use the margin loss as the global loss function. It can be defined as follows:
\begin{equation}
\label{eq:lossF}
\begin{split}
{L}=\sum_{k=1}^{n}({T_{k}}&{\rm{max}}(0,r^{+}-\Vert \textbf{u}_{k}^{(4)}\Vert)^{2}\\
&+\lambda (1-{T_{k}}){\rm{max}}(0,\Vert \textbf{a}_{k}^{(4)}\Vert-r^{-})^{2})
\end{split}
\end{equation}
where $T_{k}$ is an indicator function. 
It can be defined by
\begin{equation}
\label{eq:indiF}
\begin{split}
{T_{k}} =
\begin{cases}
1,  & \text{if class $k$ is present in sample} \\
0, & \text{otherwise}.
\end{cases}
\end{split}
\end{equation}
Herein, the indicator function $T_{k}$ is used to indicate which part (addends) of $L$ is active. 
The first part works when $T_{k}=1$. 
Otherwise, the second part works when $T_{k}=0$. 
In order to avoid the maximum or collapse loss, the loss function $L$ introduces the concept of boundary, forcing the length of the activity vector $\Vert \textbf{u}_{k}^{(4)}\Vert$ falls into small interval. 
The boundary parameters $r^{+}$ and $r^{-}$ are upper boundary and lower boundary, respectively.  
Additionally, the regularization parameter $\lambda$ is used to shrink the influence of activity vector when the corresponding class does not exist in sample.

\section{Experimental Results}
\label{sec:05}
In this section, we evaluate the performance of our proposed method on three representative HSI datasets.
Firstly, we introduce three hyperspectral datasets, which are used to verify our proposed \textit{1D-ConvCapsNet}. 
Secondly, we provide the hyperparameters setup, experimental environment and contrastive methods. 
Finally, we compare experimental results and give relative discussions.

\begin{figure}[!h]
	\centering
	\includegraphics{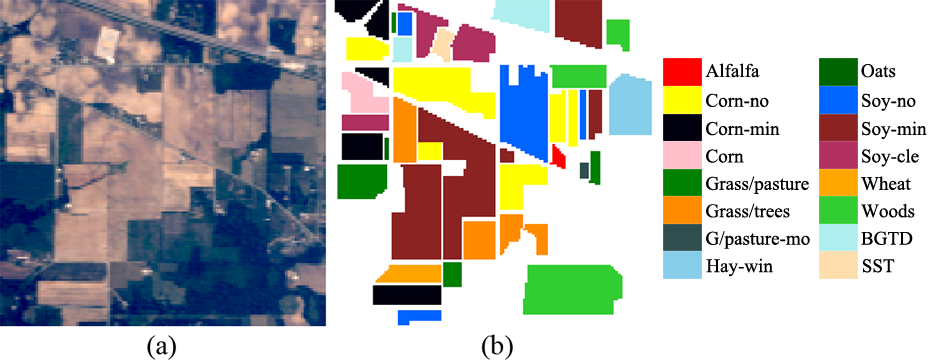}
	\caption{The real image (a) and ground truths map (b) of Indian Pines dataset.}
	\label{fig:datasetIP} 
\end{figure}
\begin{table}[!h]
	\centering
	\caption{The Sample Numbers of 20\% Training Set, 10\% Verification Set and 70\% Test Set on Indian Pines Dataset}
	\label{tab:datasetIP} 
	\begin{tabular}{ccccc}
		\hline\noalign{\smallskip}
		Label & Class & Training & Validation & Test\\
		\hline\noalign{\smallskip}
		1 & Alfalfa & 9 & 4 & 33\\
		2 & Corn-no & 285 & 142 & 1,001\\
		3 & Corn-min & 166 & 83 & 581\\
		4 & Corn & 47 & 23 & 167\\
		5 & Grass/pasture & 96 & 48 & 339\\
		6 & Grass/trees & 146 & 73 & 511\\
		7 & Grass/pasture-mo & 5 & 2 & 21\\
		8 & Hay-win & 95 & 47 & 336\\
		9 & Oats & 4 & 2 & 14\\
		10 & Soy-no & 194 & 97 & 681\\
		11 & Soy-min & 491 & 245 & 1,719\\
		12 & Soy-cle & 118 & 59 & 416\\
		13 & Wheat & 41 & 20 & 144\\
		14 & Woods & 253 & 126 & 886\\
		15 & BGTD & 77 & 38 & 271\\
		16 & SST & 18 & 9 & 66\\
		Total & $--$ & 2,045 & 1,018 & 7,186\\
		\hline\noalign{\smallskip}
	\end{tabular}
\end{table}
\begin{figure}[!h]
	\centering
	\includegraphics{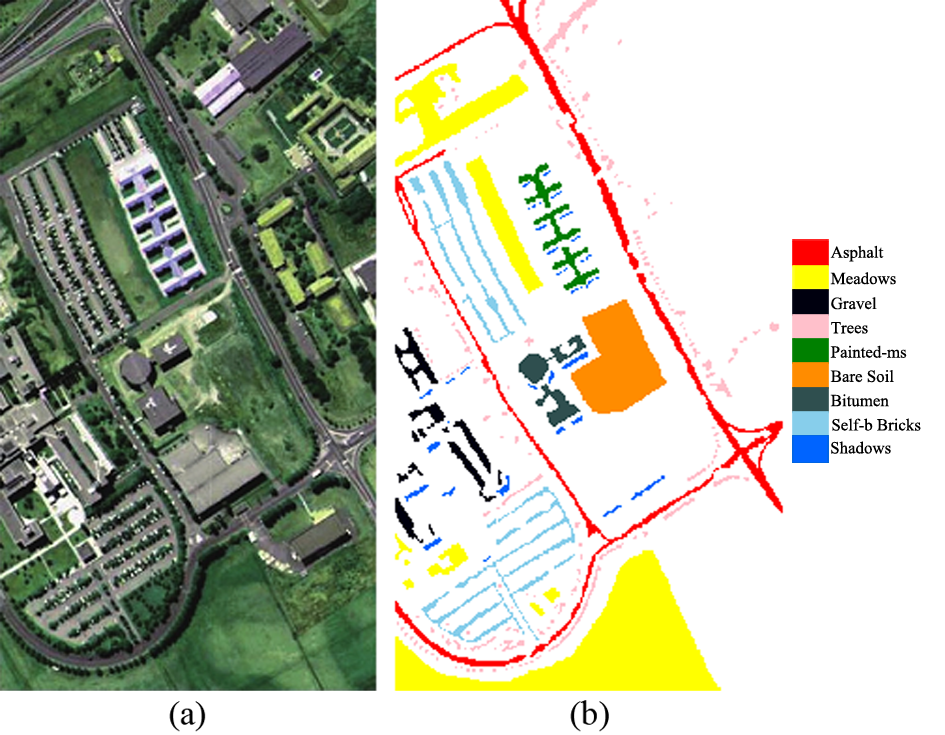}
	\caption{The real image (a) and ground truths map (b) of University of Pavia.}
	\label{fig:datasetUP} 
\end{figure}
\begin{table}[!h]
	\centering
	\caption{The Sample Numbers of 20\% Training Set, 10\% Verification Set and 70\% Test Set on University of Pavia Dataset}
	\label{tab:datasetUP}
	\begin{tabular}{cccccc}
		\hline\noalign{\smallskip}
		Label & Class & Training & Validation & Test\\
		\hline\noalign{\smallskip}
		1 & Asphalt & 1,326 & 663 & 4,642\\
		2 & Meadows & 3,729 & 1864 &13,056\\
		3 & Gravel & 419 & 209 & 1,471\\
		4 & Trees & 612 & 306 & 2,146\\
		5 & Painted-ms & 269 & 134 & 942\\
		6 & Bare Soil & 1,005 & 502 & 3,522\\
		7 & Bitumen & 266 & 133 & 931\\
		8 & Self-b Bricks & 736 & 368 &2,578\\
		9 & Shadows & 189 & 94 & 664\\
		Total & $--$ & 8,551 & 4,273 & 29,952\\
		\hline\noalign{\smallskip}
	\end{tabular}
\end{table}
\begin{figure}[!h]
	\centering
	\includegraphics{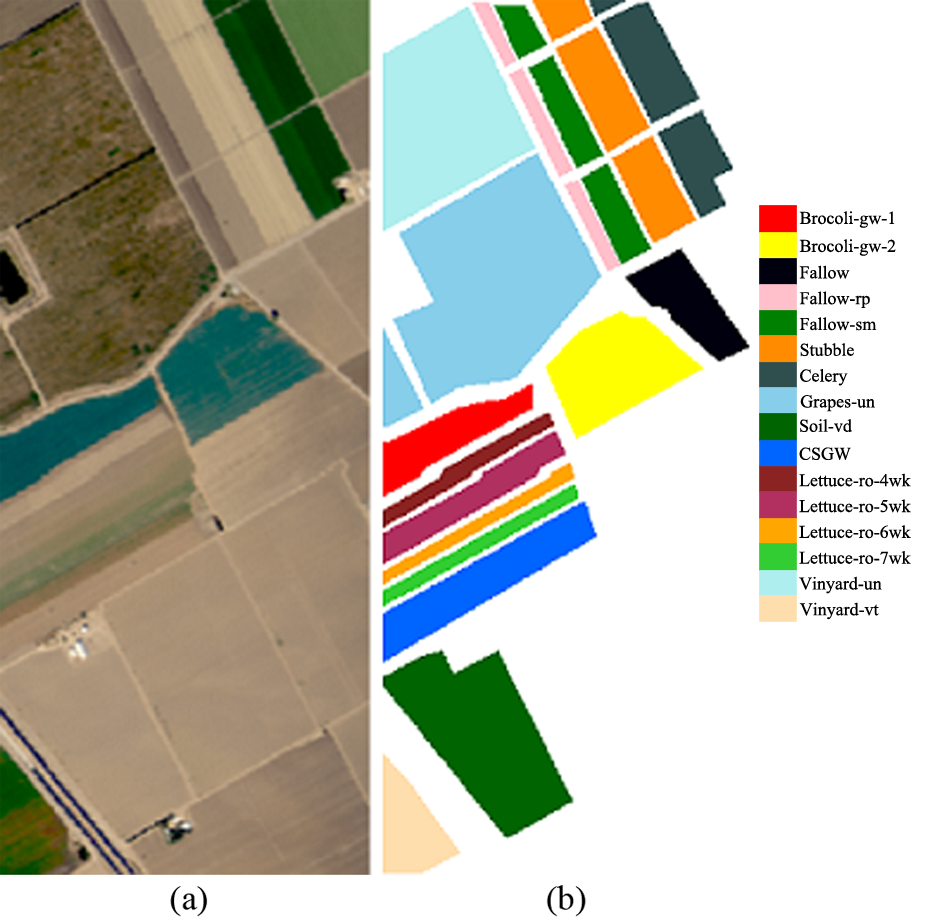}
	\caption{The real image (a) and ground truths map (b) of Salinas.}
	\label{fig:datasetSA} 
\end{figure}
\begin{table}[!h]
	\centering
	\caption{The Sample Numbers of 20\% Training Set, 10\% Verification Set and 70\% Test Set on Salinas Dataset}
	\label{tab:datasetSA} 
	\begin{tabular}{ccccc}
		\hline\noalign{\smallskip}
		Label & Class & Training & Validation & Test\\
		\hline\noalign{\smallskip}
		1 & Brocoli-gw-1 & 401 & 200 & 1,408\\
		2 & Brocoli-gw-2 & 745 & 372 & 2,609\\
		3 & Fallow & 395 & 197 & 1,384\\
		4 & Fallow-rp & 278 & 139 & 977\\
		5 & Fallow-sm & 535 & 267 & 1,876\\
		6 & Stubble & 791 & 395 & 2,773\\
		7 & Celery & 715 & 357 & 2,507\\
		8 & Grapes-un & 2,254 & 1,127 & 7,890\\
		9 & Soil-vd & 1,240 & 620 & 4,343\\
		10 & CSGW & 655 & 327 & 2,296\\
		11 & Lettuce-ro-4wk & 213 & 106 & 749\\
		12 & Lettuce-ro-5wk & 385 & 192 & 1,350\\
		13 & Lettuce-ro-6wk & 183 & 91 & 642\\
		14 & Lettuce-ro-7wk & 214 & 107 & 749\\
		15 & Vinyard-un & 1,453 & 726 & 5,089\\
		16 & Vinyard-vt & 361 & 180 & 1,266\\
		Total & $--$ & 10,818 & 5,403 & 37,908\\
		\hline\noalign{\smallskip}
	\end{tabular}
\end{table}
\subsection{Hyperspectral Datasets and Preprocessing}
\label{sec:051}
Three representative hyperspectral datasets in HSIs classification are selected, which are Indian Pines (IP), University of Pavia (UP) and Salinas (SA), respectively. 
IP has uneven distribution of sample numbers and low spatial resolution, UP has small spectral channels, and SA has large data volume. 
These characteristics are used to verify that whether \textit{1D-ConvCapsNet} can achieve the desired performance under a variety of conditions. 
\subsubsection{Indian Pines}
\label{sec:0511}
IP scene is the subset of Big Indian Pines (BIP) scene, which is generated by the Airborne Visible/Infrared Imaging Spectrometer (AVIRIS) sensor at Northwestern Indiana, USA, in 1992. 
It consists of several agricultural crops, such as corn, oat and wheat. 
The IP contains $145\times145$ pixels with a spatial resolution of 20m, including $16$ classes of interested ground truths. 
Each pixel consists of 220 SCs, which cover wavelengths from 400 to 2500nm. 
The real image and label map are shown in Fig. \ref{fig:datasetIP}. 
The numbers of each ground truth in training, validation and test sets are recorded in Table \ref{tab:datasetIP}. 
We can see that there are a few samples of some ground truths, i.e., Alfalfa, Grass/pasture-mo and Oats. 
\subsubsection{University of Pavia}
\label{sec:0512}
UP scene is generated by the Reflective Optics System Imaging Spectrometer (ROSIS) sensor in the city of Pavia, Italy, during a flight campaign over Pavia in 2001. 
It is a typical image of the city, including many building materials, such as bricks, asphalt and metal sheets. 
The UP contains $610\times 340$ pixels with a spatial resolution of 1.3m, including $9$ classes of interested ground truths. 
Each pixel consists of 103 SCs, which cover wavelengths from 430 to 860nm. 
The real image and label map are shown in Fig. \ref{fig:datasetUP}. 
The numbers of each ground truth in training, validation and test sets are recorded in Table \ref{tab:datasetUP}. 
\subsubsection{Salinas}
\label{sec:0513}
SA scene is generated by AVIRIS sensor at Salinas valley, California, USA, in 1992. 
Like IP, it also consists of several agriculture related fields, such as vegetables, bare soils, and vineyard fields. 
The SA contains $512\times 217$ pixels with a spatial resolution of 3.7m, including $16$ classes of interested ground truths. 
Each pixel consists of 224 SCs, which cover wavelengths from 400 to 2500nm. 
The real image and label map are shown in Fig. \ref{fig:datasetSA}. 
The number of each ground truth in training, validation and test sets are recorded in Table \ref{tab:datasetSA}.
\subsubsection{Data Preprocessing}
\label{sec:0514}
The performance of classifiers is affected by using raw data due to the high correlations between SCs. 
Hence, it is necessary to preprocess the data before going to the training stage.
In our method, the correlation between SCs is eliminated by using PCA-Whitening.

\subsection{Exprimential Setup}
\label{sec:052}
\begin{table}
	\centering
	\caption{The Structure Setup of 1D-ConvCapsNet on Three Datasets.}
	\label{tab:hypa}
	\newcommand{\tabincell}[2]{\begin{tabular}{@{}#1@{}}#2\end{tabular}}
	\renewcommand{\arraystretch}{1.25}
	\begin{tabular}{|p{2.5cm}<{\centering}|p{2.5cm}<{\centering}|p{2.5cm}<{\centering}|}
		\hline
		\multicolumn{3}{|c|}{\textbf{SpatialConv Layer}} \\
		\hline
		$Layer\;ID$ & $Input\;size$ & $f^{(1)}@\mathcal{K}^{(1)}$ \\
		\hline
		$\mathcal{L}^{(1)}$ & $(7\times 7\times \mathcal{C})$ & $(7\times 7)@16$ \\
		\hline
		$Stride$ & $Activation\;function$ & $Output\;size$ \\
		\hline
		$(1,1)$ & $ReLU$ & $(\mathcal{C}\times 16)$ \\
		\hline
		\multicolumn{3}{|c|}{\textbf{PrimaryCaps Layer}} \\
		\hline
		$Layer\;ID$ & $Input\;size$ & $f^{(2)}@\mathcal{K}^{(2)}$ \\
		\hline
		$\mathcal{L}^{(2)}$ & $(\mathcal{C}\times 16)$ & $(9)@16$ \\
		\hline
		$Stride$ & $Activation\;function$ & $Output\;size$ \\
		\hline
		$(2)$ & $ReLU$ & $(c^{(2)}\times2 \times 8)$ \\
		\hline
		\multicolumn{3}{|c|}{\textbf{1D-ConvCaps Layer}} \\
		\hline
		$Layer\;ID$ & $Input\;size$ & $f^{(3)}@\mathcal{K}^{(3)}$ \\
		\hline
		$\mathcal{L}^{(3)}$ & $(c^{(2)}\times2 \times 8)$ & $(9)@4$ \\
		\hline
		$Stride$ & $Activation\;function$ & $Output\;size$ \\
		\hline
		$(2)$ & $squashing$ & $(c^{(3)}\times4 \times 8)$ \\
		\hline
		\multicolumn{3}{|c|}{\textbf{ClassCaps Layer}} \\
		\hline
		$Layer\;ID$ & $Input\;size$ & $Filters\;size$ \\
		\hline
		$\mathcal{L}^{(4)}$ & $(c^{(3)}\times4 \times 8)$ & $--$ \\
		\hline
		$Stride$ & $Activation\;function$ & $Output\;size$ \\
		\hline
		$--$ & $squashing$ & $(n\times 16)$ \\
		\hline
	\end{tabular}
\end{table}
\subsubsection{Hyperparameters}
\label{sec:0521}
The optimal structure of \textit{1D-ConvCapsNet} is determined by repeated experiments, which is shown in Table \ref{tab:hypa}. 
Adam optimizer is used to train 50 epochs with 0.01 learning rate and 3 routing iterations without regularization and normalization. 
The batch size is set to 64 for the IP and 256 for the rest, which is related to the number of samples. 
The Hyperparameters $r^{+}$, $r^{-}$ and $\lambda$ in the loss function are set to 0.9, 0.1 and 0.5, respectively. 
During the experiments, 20\% patches of each class are randomly selected as the training set, 10\% patches as the validation set and the rest patches as the test set. 
The accuracy on the validation set is recorded during every epoch, the highest of which is regard as optimal parameters to verify the performance on test set.

\subsubsection{Experimental Environment}
\label{sec:0522}
All experiments are conducted under the same environment. 
The hardware platform consists of Intel Core i7-7820HK processor (four core/eight threading) with 8M L3-cache, 16GB DDR4 memory with 2800Mhz serial speed, Nvidia GeForce GTX 1070 GPU with 8GB DDR5 video memory and 1TB HDD with 7200 RPM. 
The software platform includes Windows 10 Professional operating system, Keras 2.1.1 based on TensorFlow-gpu 1.3.0 and Python 3.5.2.

\subsubsection{Comparative methods}
\label{sec:0523}
In experiment 1, four well-known HSIs classification methods are selected as comparative methods. 
They are SVM with radial basis function (RBF-SVM)\cite{IEEEhowto:5_9_1}, MLP with four hidden layers, semi-supervised convolutional neural network (Se-2D-CNN)\cite{IEEEhowto:2_1_10} and 3D-convolutional neural network (3D-CNN)\cite{IEEEhowto:1_3_7}. 
Note that RBF-SVM and MLP focus on spectral information, Se-2D-CNN focuses on spatial information, and 3D-CNN and \textit{1D-ConvCapsNet} consider spectral-spatial information for HSIs classification. In Experiment 2, CNN model and CapsNet\cite{IEEEhowto:1_5_1} are selected as comparative methods to compare the gaps between different training methods. 
The structure of CNN model is similar to \textit{1D-ConvCapsNet} except the neuron type.

In order to quantify the accuracy of classifiers, overall accuracy (OA), average classification accuracy (AA) and kappa coefficient ($k$) serve as evaluation metrics. 
In order to obtain stable results, we conduct 20 experiments and take the median of OA as convincing result. 
The training time is also recorded to evaluate the time cost in the training stage. 
Since the training time is affected by the system utilization rate and other factors, the minimal training time in 20 experiments is recorded as the result.

\subsection{Results of experiment and Discussion}
\label{sec:053}
\subsubsection{Experiments 1}
\label{sec:0531}
\begin{table}
	\centering
	\caption{Classification Results Obtained by Different Mehtods on Indian Pines Dataset}
	\label{tab:resultIP1}
	\newcommand{\tabincell}[2]{\begin{tabular}{@{}#1@{}}#2\end{tabular}}
	\begin{tabular}{cccccc}
		\hline\noalign{\smallskip}
		Class & RBF-SVM & MLP & Se-2D-CNN & 3D-CNN & Proposed\\
		\hline\noalign{\smallskip}
		1 & 36.96\% & 56.25\% & 0\% & \textbf{100\%} & \textbf{100\%} \\
		2 & 81.58\% & 77.50\% & 96.60\% & 97.90\% & \textbf{99.60\%} \\
		3 & 73.37\% & 67.81\% & 89.16\% & 96.39\% & \textbf{99.14\%} \\
		4 & 65.40\% & 67.47\% & 54.22\% & 95.78\% & \textbf{98.20\%} \\
		5 & 89.44\% & 94.97\% & 92.90\% & \textbf{97.63\%} & 97.35\% \\
		6 & 97.12\% & 98.04\% & 99.41\% & 99.02\% & \textbf{99.80\%} \\
		7 & 42.86\% & 95.00\% & 0\% & 85\% & \textbf{100\%} \\
		8 & 98.33\% & 99.40\% & 100\% & 98.81\% & \textbf{100\%} \\
		9 & 35\% & 57.14\% & 0\% & 85.71\% & \textbf{100\%} \\
		10 & 77.06\% & 82.50\% & 91.18\% & 96.62\% & \textbf{99.12\%} \\
		11 & 85.01\% & 91.45\% & 98.84\% & 99.53\% & \textbf{99.71\%} \\
		12 & 76.05\% & 88.67\% & 96.14\% & 97.35\% & \textbf{97.84\%} \\
		13 & 96.10\% & 98.60\% & 96.50\% & 98.60\% & \textbf{100\%} \\
		14 & 95.18\% & 94.92\% & 99.10\% & 99.21\% & \textbf{99.77\%} \\
		15 & 68.13\% & 69.63\% & 94.44\% & \textbf{100\%} & 95.57\% \\
		16 & 92.47\% & 90.77\% & 67.69\% & 81.54\% & \textbf{95.45\%} \\
		OA & 84.04\% & 86.56\% & 94.27\% & 98.13\% & \textbf{99.18\%} \\
		AA & 75.63\% & 83.13\% & 73.51\% & 95.57\% & \textbf{98.85\%} \\
		$k\times 100$ & 81.75 & 84.62 & 93.44 & 97.87 & \textbf{99.06} \\
		\hline\noalign{\smallskip}
		Time & 5s & 317s & 424s & 18,672s & 402s \\
		\hline
	\end{tabular}
\end{table}
\begin{table}
	\centering
	\caption{Classification Results Obtained by Different Mehtods on University of Pavia Dataset}
	\label{tab:resultUP1} 
	\begin{tabular}{cccccc}
		\hline\noalign{\smallskip}
		Class & RBF-SVM & MLP & Se-2D-CNN & 3D-CNN & Proposed\\
		\hline\noalign{\smallskip}
		1 & 94.71\% & 96.06\% & 99.27\% & \textbf{99.85\%} & 99.70\% \\
		2 & 98.48\% & 97.02\% & 99.89\% & 99.92\% & \textbf{99.95\%} \\
		3 & 79.28\% & 87.20\% & 86.79\% & 97.82\% & \textbf{98.37\%} \\
		4 & 95.53\% & 96.46\% & 98.14\% & 98.55\% & \textbf{99.44\%} \\
		5 & 99.78\% & 99.58\% & \textbf{100\%} & \textbf{100\%} & \textbf{100\%} \\
		6 & 88.82\% & 94.55\% & 99.86\% & 98.21\% & \textbf{100\%} \\
		7 & 88.05\% & 90.98\% & 94.52\% & 99.46\% & \textbf{99.89\%} \\
		8 & 91.69\% & 90.73\% & 98.29\% & \textbf{99.26\%} & 98.33\% \\
		9 & \textbf{100\%} & 99.40\% & 99.85\% & 98.34\% & 99.85\% \\
		OA & 94.77\% & 95.46\% & 98.72\% & 99.40\% & \textbf{99.66\%} \\
		AA & 92.92\% & 94.66\% & 97.40\% & 99.05\% & \textbf{99.50\%} \\
		$k\times 100$ & 93.04 & 94.00 & 98.30 & 99.21 & \textbf{99.55} \\
		\hline\noalign{\smallskip}
		Time & 11s & 1,185s & 3,618s & 30,363s & 432s \\
		\hline
	\end{tabular}
\end{table}
\begin{table}
	\centering
	\caption{Classification Results Obtained by Different Mehtods on Salinas Dataset}
	\label{tab:resultSA1} 
	\begin{tabular}{cccccc}
		\hline\noalign{\smallskip}
		Class & RBF-SVM & MLP & Se-2D-CNN & 3D-CNN & Proposed\\
		\hline\noalign{\smallskip}
		1 & 99.90\% & 99.57\% & 99.22\% & \textbf{100\%} & \textbf{100\%} \\
		2 & \textbf{100\%} & 99.88\% & \textbf{100\%} & \textbf{100\%} & \textbf{100\%} \\
		3 & 99.70\% & 99.28\% & 98.05\% & 99.86\% & \textbf{100\%} \\
		4 & 99.28\% & 99.69\% & 99.80\% & 99.59\% & \textbf{100\%} \\
		5 & 99.22\% & 99.68\% & 99.09\% & \textbf{100\%} & 99.89\% \\
		6 & 99.90\% & 99.93\% & \textbf{100\%} & \textbf{100\%} & \textbf{100\%} \\
		7 & 99.94\% & 99.84\% & 99.96\% & \textbf{100\%} & \textbf{100\%} \\
		8 & 90.10\% & 90.44\% & 92.86\% & \textbf{99.96\%} & 99.82\% \\
		9 & 99.95\% & 99.95\% & 99.91\% & \textbf{100\%} & \textbf{100\%} \\
		10 & 97.35\% & 97.47\% & 97.12\% & \textbf{99.97\%} & 99.96\% \\
		11 & 95.13\% & 98.26\% & 99.47\% & 98.66\% & \textbf{100\%} \\
		12 & 99.84\% & 99.93\% & \textbf{100\%} & \textbf{100\%} & \textbf{100\%} \\
		13 & 99.34\% & 98.75\% & 99.06\% & \textbf{100\%} & \textbf{100\%} \\
		14 & 96.92\% & 98.53\% & 98.80\% & \textbf{100\%} & 99.33\% \\
		15 & 72.30\% & 82.29\% & 76.14\% & 99.27\% & \textbf{99.57\%} \\
		16 & 99.39\% & 99.37\% & 99.13\% & \textbf{100\%} & 99.92\% \\
		OA & 93.78\% & 95.28\% & 94.89\% & 99.85\% & \textbf{99.88\%} \\
		AA & 96.77\% & 97.68\% & 97.41\% & 99.83\% & \textbf{99.91\%} \\
		$k\times 100$ & 94.90 & 94.74 & 94.31 & 99.83 & \textbf{99.87} \\
		\hline\noalign{\smallskip}
		Time & 48s & 1,525s & 2,284s & 101,873s & 2,146s \\
		\hline
	\end{tabular}
\end{table}
\begin{figure*}
	\centering
	\includegraphics{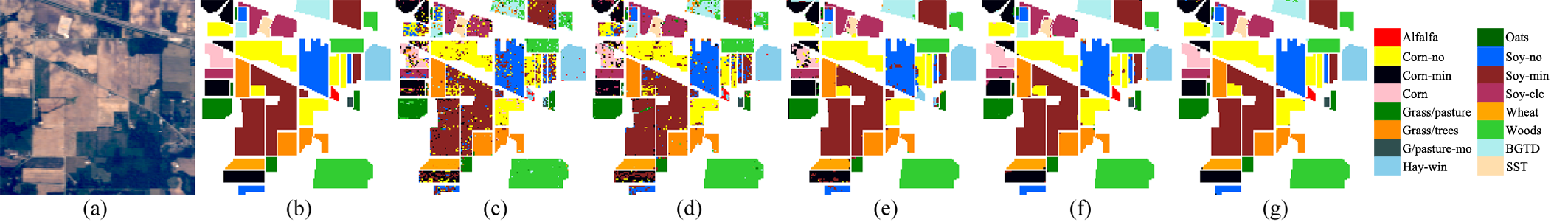}
	\caption{Classification maps obtained by different mehtods on Indian Pines dataset. (a) Real image. (b) Classification map. (c) RBF-SVM. (d) MLP. (e) Se-2D-CNN. (f) 3D-CNN. (g) 1D-ConvCapsNet.}
	\label{fig:resultIP1} 
\end{figure*}
\begin{figure*}
	\centering
	\includegraphics{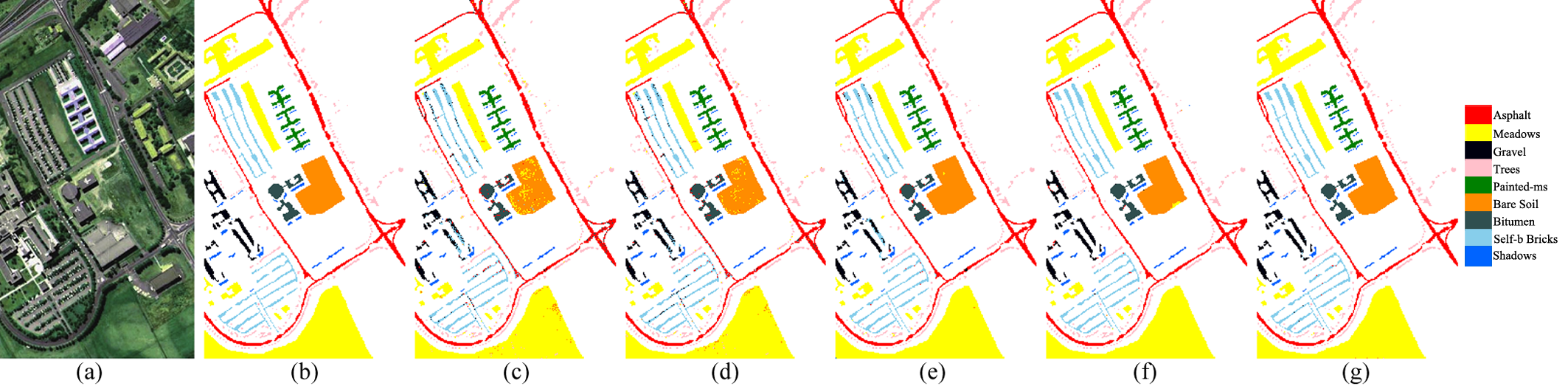}
	\caption{Classification maps obtained by different mehtods on University of Pavia dataset. (a) Real image. (b) Classification map. (c) RBF-SVM. (d) MLP. (e) Se-2D-CNN. (f) 3D-CNN. (g) 1D-ConvCapsNet.}
	\label{fig:resultUP1} 
\end{figure*}
\begin{figure*}
	\centering
	\includegraphics{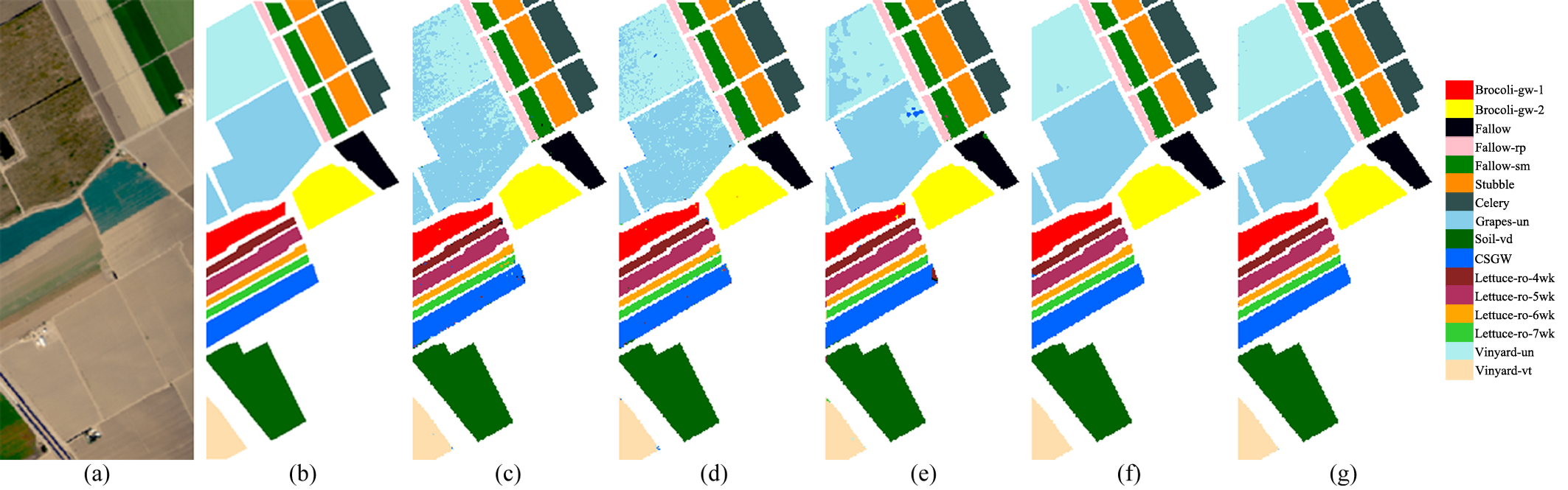}
	\caption{Classification maps obtained by different mehtods on Salinas dataset. (a) Real image. (b) Classification map. (c) RBF-SVM. (d) MLP. (e) Se-2D-CNN. (f) 3D-CNN. (g) 1D-ConvCapsNet.}
	\label{fig:resultSA1}  
\end{figure*}
This experiment validate the performances of the proposed and comparative methods on IP, UP and SA. 
In the RBF-SVM, the parameters $\gamma$ and $C$ are set to $0.005$ and $100$, respectively. 
In the MLP, the number of neurons with $ReLU$ activation function and 50\% dropout probability for each layer is 2048, 4096, 2048 and 2048, respectively, which uses Adam optimizer to train 500 epochs with 0.01 learning rate. 
In the Se-2D-CNN, the default settings are used. 
In the 3D-CNN, due to the extremely large model scale, it cannot be directly run in our environment. 
In order to make 3D-CNN work, the number of filters in every layer is reduced to 32, and the rest of settings are unchanged. 
Table \ref{tab:resultIP1}-\ref{tab:resultSA1} records the accuracy and the evaluation metrics of all methods on IP, UP and SA, where each row represents the classification accuracy and each column represents each method.

From the results recorded in Table \ref{tab:resultIP1}-\ref{tab:resultSA1}, we can observe that the accuracy of the proposed method is superior to the RBF-SVM, MLP, Se-2D-CNN and 3D-CNN on three datasets. 
Compared to 3D-CNN, the accuracy of \textit{1D-ConvCapsNet} on IP has great improvement, but on UP and SA it is not so obvious. 
This is because the proposed method does not adjust structure and hyperparameters for UP and SA. 
Generally, the accuracy of \textit{1D-ConvCapsNet} and 3D-CNN are superior to the rest of comparative methods. 
This result is expected because RBF-SVM and MLP only focus on spectral information, and Se-2D-CNN only focus on spatial information. 
Our proposed method and 3D-CNN can extract information from spectral and spatial domains, leading to higher accuracy. 
However, \textit{1D-ConvCapsNet} only trains 50 epochs, which means that our method converges faster than 3D-CNN and has lower training cost. 
The training time of \textit{1D-ConvCapsNet} on three datasets is about 2\% of 3D-CNN. 
Moreover, \textit{1D-ConvCapsNet} can achieve outstanding performance on classes with small number of samples, i.e., Alfalfa, Grass/pasture-mo, Oats and SST of IP. 
We can see from the Table \ref{tab:resultIP1}, all comparative methods perform unsatisfactorily on those classes, because these methods cannot generalize the entire ground truths using very few training samples. 
Especially for Se-2D-CNN, those classes do not provide enough texture information, resulting in poor accuracy performance. 
The performance of \textit{1D-ConvCapsNet} on those classes is much better than comparative methods, which benefit from powerful representation and interpretation capabilities of the capsule network.

In order to intuitively represent the performance of different methods, all samples are input into the trained classifiers to obtain classification map for each method, which are shown in Fig. \ref{fig:resultIP1}-\ref{fig:resultSA1}. 
It can be seen that the classification maps of SVM and MLP contain a lot of noise. 
This is because the quality of spectral information is affected by spatial resolution and imaging conditions, resulting in the phenomenon that the same spectrum expresses different objects or same objects have different spectrum. 
Therefore, classifiers which only focus on spectral information have high error rates.
The classification map of Se-2D-CNN contains less noise than that of SVM and MLP.  
However, Se-2D-CNN does not perform well on the edges of classes and ground truths with similar textures, because these samples provide insufficient spatial information for discrimination. 
3D-CNN and \textit{1D-ConvCapsNet} achieve higher accuracy and less noise than other comparative methods, because they can act on both spatial information and spectral information.
This result is consistent with Table \ref{tab:resultIP1}-\ref{tab:resultSA1}.

\begin{table}
	\centering
	\caption{Accuracy of c-CNN, CapsNet and Our Method on Indian Pines Dataset}
	\label{tab:resultIP2}
	\newcommand{\tabincell}[2]{\begin{tabular}{@{}#1@{}}#2\end{tabular}}
	\begin{tabular}{ccccc}
		\hline\noalign{\smallskip}
		Label & Class & c-CNN & CapsNet & Proposed \\
		\hline\noalign{\smallskip}
		1 & Alfalfa & \textbf{100\%} & 96.97\% & \textbf{100\%} \\
		2 & Corn-no & 98.50\% & 96.50\% & \textbf{99.60\%} \\
		3 & Corn-min & 92.60\% & 98.62\% & \textbf{99.14\%} \\
		4 & Corn & 97.01\% & \textbf{98.20\%} & \textbf{98.20\%} \\
		5 & Grass/pasture & 94.40\% & 94.69\% & \textbf{97.35\%} \\
		6 & Grass/trees & 99.22\% & 99.61\% & \textbf{99.80\%} \\
		7 & Grass/pasture-no & 90.48\% & \textbf{100\%} & \textbf{100\%} \\
		8 & Hay-win & 99.40\% & \textbf{100\%} & \textbf{100\%} \\
		9 & Oats & 92.86\% & \textbf{100\%} & \textbf{100\%} \\
		10 & Soy-no & 95.59\% & 98.38\% & \textbf{99.12\%} \\
		11 & Soy-min & 96.92\% & \textbf{97.85\%} & 97.71\% \\
		12 & Soy-cle & 97.60\% & 95.91\% & \textbf{97.84\%} \\
		13 & Wheat & \textbf{100\%} & \textbf{100\%} & \textbf{100\%} \\
		14 & Woods & 99.66\% & 99.66\% & \textbf{99.77\%} \\
		15 & BGTD & 94.83\% & 92.99\% & \textbf{95.57\%} \\
		16 & SST & 90.91\% & 86.36\% & \textbf{95.45\%} \\
		OA & $--$ & 97.12\% & 97.73\% & \textbf{99.18\%} \\
		AA & $--$ & 96.25\% & 97.23\% & \textbf{98.85\%} \\
		$k\times 100$ & $--$ & 96.71 & 97.41 & \textbf{99.06} \\
		\hline
	\end{tabular}
\end{table}
\begin{table}
	\centering
	\caption{Accuracy of c-CNN, CapsNet and Our Method on University of Pavia Dataset}
	\label{tab:resultUP2}
	\begin{tabular}{ccccc}
		\hline\noalign{\smallskip}
		Label & Category & c-CNN & CapsNet & Proposed\\
		\hline\noalign{\smallskip}
		1 & Asphalt & 99.12\% & 99.35\% & \textbf{99.70\%} \\
		2 & Meadows & 99.83\% & \textbf{100\%} & 99.95\% \\
		3 & Gravel & 94.83\% & 94.02\% & \textbf{98.37\%} \\
		4 & Trees & 98.70\% & 99.21\% & \textbf{99.44\%} \\
		5 & Painted-ms & 99.26\% & \textbf{100\%} & \textbf{100\%} \\
		6 & Bare Soil & 99.38\% & 99.97\% & \textbf{100\%} \\
		7 & Bitumen & 98.60\% & 99.68\% & \textbf{99.89\%} \\
		8 & Self-b Bricks & 95.31\% & 97.91\% & \textbf{98.33\%} \\
		9 & Shadows & 96.23\% & \textbf{99.85\%} & \textbf{99.85\%} \\
		OA & $--$ & 98.81\% & 99.35\% & \textbf{99.66\%} \\
		AA & $--$ & 97.91\% & 98.89\% & \textbf{99.50\%} \\
		$k\times 100$ & $--$ & 98.43 & 99.14 & \textbf{99.55} \\
		\hline
	\end{tabular}
\end{table}
\begin{table}
	\centering
	\caption{Accuracy of c-CNN, CapsNet and Our Method on Salinas Dataset}
	\label{tab:resultSA2}
	\begin{tabular}{ccccc}
		\hline\noalign{\smallskip}
		Label & Category & c-CNN & CapsNet & Proposed \\
		\hline\noalign{\smallskip}
		1 & Brocoli-gw-1 & 99.72\% & \textbf{100\%} & \textbf{100\%} \\
		2 & Brocoli-gw-2 & \textbf{100\%} & \textbf{100\%} & \textbf{100\%} \\
		3 & Fallow & \textbf{100\%} & \textbf{100\%} & \textbf{100\%} \\
		4 & Fallow-rp & 99.80\% & \textbf{100\%} & \textbf{100\%} \\
		5 & Fallow-sm & 99.52\% & \textbf{99.89\%} & \textbf{99.89\%} \\
		6 & Stubble & 99.89\% & \textbf{100\%} & \textbf{100\%} \\
		7 & Celery & \textbf{100\%} & \textbf{100\%} & \textbf{100\%} \\
		8 & Grapes-un & 98.68\% & 99.70\% & \textbf{99.82\%} \\
		9 & Soil-vd & \textbf{100\%} & \textbf{100\%} & \textbf{100\%} \\
		10 & CSGW & 99.61\% & 99.78\% & \textbf{99.96\%} \\
		11 & Lettuce-ro-4wk & 99.33\% & \textbf{100\%} & \textbf{100\%} \\
		12 & Lettuce-ro-5wk & 99.78\% & \textbf{100\%} & \textbf{100\%} \\
		13 & Lettuce-ro-6wk & 98.29\% & 99.69\% & \textbf{100\%} \\
		14 & Lettuce-ro-7wk & 98.13\% & \textbf{99.60\%} & 99.33\% \\
		15 & Vinyard-un & 97.68\% & \textbf{99.65\%} & 99.57\% \\
		16 & Vinyard-vt & 99.45\% & 99.61\% & \textbf{99.92\%} \\
		OA & $--$ & 99.24\% & 99.84\% & \textbf{99.88\%} \\
		AA & $--$ & 99.37\% & 99.87\% & \textbf{99.91\%} \\
		$k\times 100$ & $--$ & 99.15 & 99.83 & \textbf{99.87} \\
		\hline
	\end{tabular}
\end{table}
\subsubsection{Experiments 2}
\label{sec:0532}
\begin{figure}
	\centering
	\includegraphics{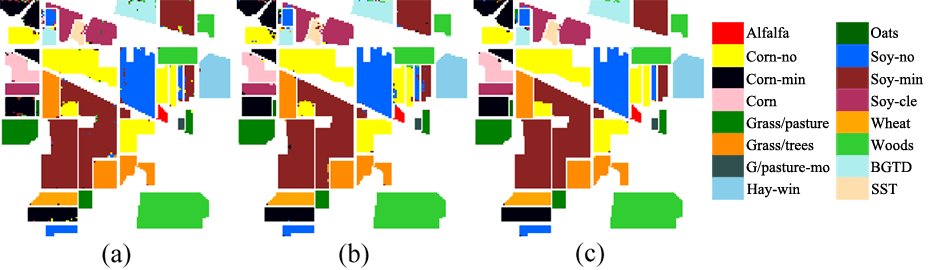}
	\caption{Classification maps obtained by (a) c-CNN, (b) CapsNet and (c) 1D-ConvCapsNet on Indian Pines dataset.}
	\label{fig:resultIP2} 
\end{figure}

\begin{figure}
	\centering
	\includegraphics{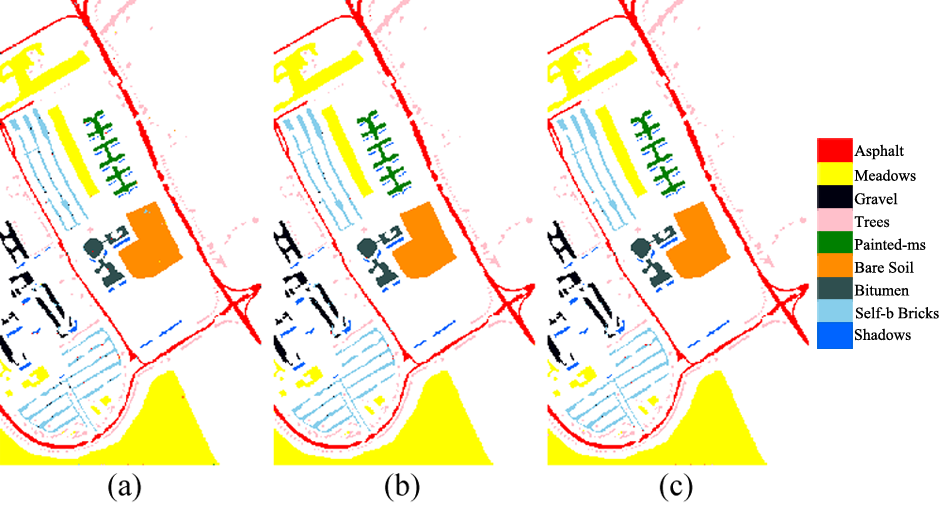}
	\caption{Classification maps obtained by (a) c-CNN, (b) CapsNet and (c) 1D-ConvCapsNet on University of Pavia dataset.}
	\label{fig:resultUP2}  
\end{figure}

\begin{figure}
	\centering
	\includegraphics{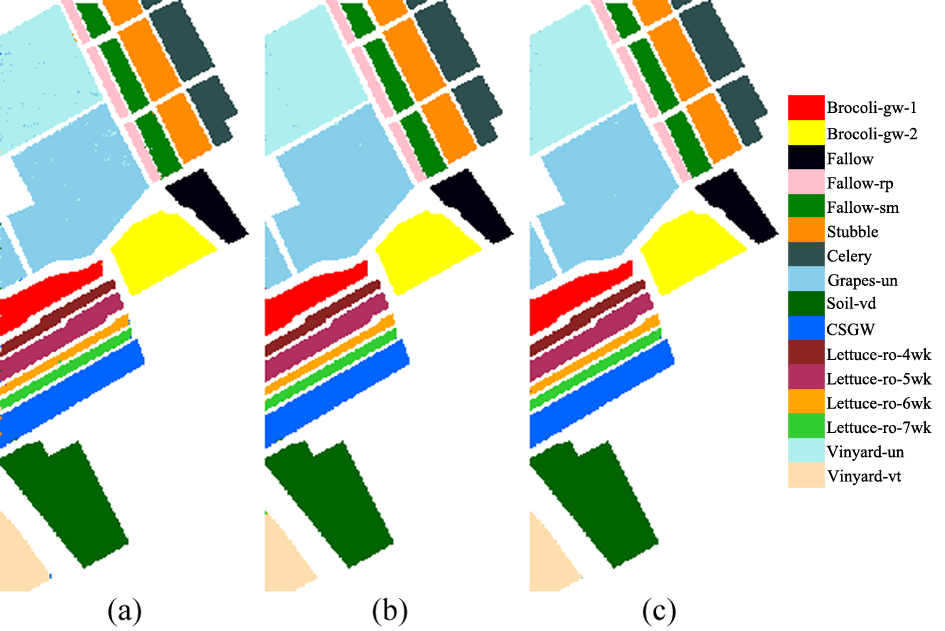}
	\caption{Classification maps obtained by (a) c-CNN, (b) CapsNet and (c) 1D-ConvCapsNet on Salinas dataset.}
	\label{fig:resultSA2} 
\end{figure}

This experiment is used to compare gaps between different training methods. 
A CNN-based model and CapsNet are selected as the comparative methods in this experiment, which extract spectral-spatial information by using our proposed methods. 
The structure of CNN-based model is identical to \textit{1D-ConvCapsNet} except the neuron type, which is denoted by c-CNN. 
In other words, all units of c-CNN are standard scalar neuron rather than vector neuron of capsule network. 
Since the original CapsNet operates on 2D-data, the 2D-convolution of CapsNet is modified to 1D-convolution after spatial information is extracted by using the SpatialConv layer. 
Table \ref{tab:resultIP2}-\ref{tab:resultSA2} records the accuracy and the evaluation metrics of three methods on IP, UP and SA, where each row represents the classification accuracy and each column represents one training method. 
Fig. \ref{fig:resultIP2}-\ref{fig:resultSA2} illustrate the classification map of each method.

From the result recorded in Table \ref{tab:resultIP2}-\ref{tab:resultSA2}, we can observe that c-CNN has significantly higher accuracy than SVM, MLP and Se-2D-CNN due to the use of our proposed spectral-spatial information extraction method. 
CapsNet and \textit{1D-ConvCapsNet} are superior to c-CNN overall, because the advantages of capsules in expressiveness and interpretability.  
It should be noted that the c-CNN achieve this result with 500 training epochs to, while the CapsNet and \textit{1D-ConvCapsNet} with 50 traing epochs. 
\textit{1D-ConvCapsNet} is superior to CapsNet in accuracy because the 1D-ConvCaps layer can learn which primary capsules are more important and combine them into complex entities.


\begin{figure*}
	\centering
	\includegraphics{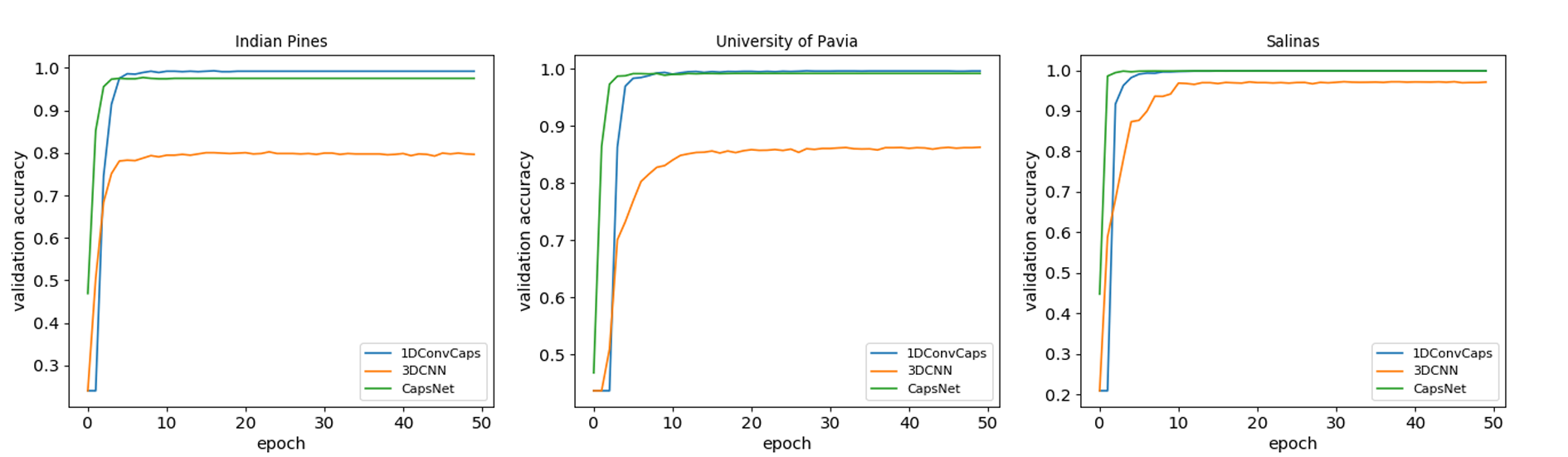}
	\caption{Convergence speed of c-CNN, CapsNet and 1D-ConvCapsNet over three datasets.}
	\label{fig:conver} 
\end{figure*}

As can be seen from Fig. \ref{fig:resultIP2}-\ref{fig:resultSA2}, classification maps obtained by CapsNet and \textit{1D-ConvCapsNet} have fewer noises than that of the c-CNN due to the high accuracy of the capsule network. 
The c-CNN has a lot of noises on Corn-no and Soy-no of IP, Bare Soil and Self-b Bricks of UP, and Grapes-un and Vinyard-un of SA. 
The \textit{1D-ConvCapsNet} have almost no errors in those classes and its classification maps are closer to label maps.
\begin{table}
	\centering
	\caption{Training Time and Parameters Numbe of 1D-ConvCapsNet and CapsNet}
	\label{tab:timeT} 
	\begin{tabular}{ccccc}
		\hline\noalign{\smallskip}
		\multirow{2}{1.0 cm}{Dataset} & \multicolumn{2}{c}{1D-ConvCapsNet} & \multicolumn{2}{c}{CapsNet} \\
		\cline{2-5}\noalign{\smallskip}
		& Time(s) & Parameters & Time(s) & Parameters \\
		\hline\noalign{\smallskip}
		Indian Pines & 402 & 409,168 & 6,483 & 7,387,904 \\
		University of Pavia & 432 & 99,920 & 6,375 & 2,325,248 \\
		Salinas & 2,146 & 417,360 & 33,723 & 7,518,976 \\
		\hline\noalign{\smallskip}
	\end{tabular}
\end{table}
\begin{table}
	\centering
	\caption{Performance of 1D-ConvCapsNet on 10\% and 5\% Training Sample.}
	\label{tab:smallS} 
	\begin{tabular}{ccc}
		\hline\noalign{\smallskip}
		\multirow{2}{1.0 cm}{Dataset} & \multicolumn{2}{c}{1D-ConvCapsNet} \\
		\cline{2-3}\noalign{\smallskip}
		& 10\% Training sample & 5\% Training sample \\
		\hline\noalign{\smallskip}
		Indian Pines & 97.89\% & 94.64\% \\
		University of Pavia & 99.28\% & 98.47\% \\
		Salinas & 99.56\% & 99.22\% \\
		\hline
	\end{tabular}
\end{table}

Fig. \ref{fig:conver} illustrates the convergence speed of three methods during the training stage, where the horizontal and vertical axes represent the number of epochs and the accuracy of verification, respectively. 
It can be seen that the two capsule-based methods converge faster than the c-CNN, and the accuracy of verification is higher than the c-CNN. 
\textit{1D-ConvCapsNet} converges slightly slower than CapsNet, but their accuracy of verification is similar. 
This is because our proposed method needs to express complex entities before connecting parts to the whole.
Table \ref{tab:timeT} records the training time and the number of parameters for CapsNet and \textit{1D-ConvCapsNet}, which represent the time cost and the storage cost, respectively. 
Compared to the CapsNet, \textit{1D-ConvCapsNet} is faster and lighter with the same accuracy, which means fewer efforts are required in the training stage. 
As can be seen in Table \ref{tab:timeT}, \textit{1D-ConvCapsNet} is about 4\%-7\% of CapsNet in terms of the training time and the number of parameters on three datasets.
This is because of the local strategy of \textit{1D-ConvCapsNet} by using local connection and sharing parameters on PrimaryCaps layer.

In order to verify the performance of \textit{1D-ConvCapsNet} on small samples, 10\% and 5\% of samples are randomly selected as the training sets in each class. 
The proportion of verification set is still 10\%, and the remaining pixels are the test set. 
The hyperparameters of \textit{1D-ConvCapsNet} are unchanged. 
The OA of \textit{1D-ConvCapsNet} on the small training sets is recorded in Table \ref{tab:smallS}.
We can see that \textit{1D-ConvCapsNet} can maintain high accuracy with 10\% training samples, and the performance by using 5\% training samples is still acceptable.

The above experimental results show that \textit{1D-ConvCapsNet} has high accuracy and low training cost.
Compared to the comparative methods, our proposed method is highly competitive. 
The accuracy of \textit{1D-ConvCapsNet} can reach the level of 3D-CNN which is state-of-the-art method introduced by\cite{IEEEhowto:1_3_7}. 
However, \textit{1D-ConvCapsNet} is much better than 3D-CNN in training speed and the number of parameters.
Compared to CapsNet, \textit{1D-ConvCapsNet} greatly reduces the number of parameters and guarantees the accuracy. 
Our methods save time and storage cost, and extend the application of capsule network.

\section{Conclusion}
\label{sec:06}
In this work, we proposed a fast and accurate capsule network for HSIs classification task, called \textit{1D-ConvCapsNet}.  
Firstly, \textit{1D-ConvCapsNet} separately extracts features on spatial and spectral domains. 
Compared to 3D-CNN, our separate feature extraction method is lightweight and fast due to fewer parameters. 
Secondly, \textit{1D-ConvCapsNet} uses local strategy to reduce the scale of capsule network. 
Finally, \textit{1D-ConvCapsNet} obtains predictions by using dynamic routing.
It is expected to have fewer parameters than several state-of-the-art methods and ensure high precision. 
The effectiveness of \textit{1D-ConvCapsNet} has been validated on three representative datasets in the HSIs classification field. 
Experimental results showed that \textit{1D-ConvCapsNet} is very competitive with the comparison algorithm. 
The accuracy of \textit{1D-ConvCapsNet} achieved the level of state-of-the-art methods, but with much lower training time and hardware requirements.
%are much lower than state-of-the-art methods. 
%
Compared to CapsNet, \textit{1D-ConvCapsNet} is about 4\%-7\% of CapsNet in terms of training time and the number of parameters on three datasets. 
\textit{1D-ConvCapsNet} also achieved outstanding performance on small samples due to powerful representation and interpretation capabilities of capsule network. 
%
%Nowadays, capsule network is a hot topic in deep learning. 
%%
%It can accomplish many challenging problems due to characteristics of capsules. 
%
In the future work, we expect to extend the \textit{1D-ConvCapsNet} to more efficient way for solving HSIs classification problem, such developing 
 more effective regularizations and the tensor constraint.
%by extending the \textit{1D-ConvCapsNet}, we expect to develop more effective regularizations on 
%to avoid overfitting. 

% if have a single appendix:
%\appendix[Proof of the Zonklar Equations]
% or
%\appendix  % for no appendix heading
% do not use \section anymore after \appendix, only \section*
% is possibly needed

% use appendices with more than one appendix
% then use \section to start each appendix
% you must declare a \section before using any
% \subsection or using \label (\appendices by itself
% starts a section numbered zero.)
%

%\appendices
%\section{Proof of the First Zonklar Equation}
%Appendix one text goes here.

% you can choose not to have a title for an appendix
% if you want by leaving the argument blank
%\section{}
%Appendix two text goes here.

% use section* for acknowledgment
%\section*{Acknowledgment}

%The authors would like to thank...

% Can use something like this to put references on a page
% by themselves when using endfloat and the captionsoff option.
\ifCLASSOPTIONcaptionsoff
  \newpage
\fi

\end{document}